\documentclass[10pt,twocolumn,letterpaper]{article}

\usepackage[pagenumbers]{cvpr} 

\usepackage{graphicx}
\usepackage{amsmath}
\usepackage{amssymb}
\usepackage{booktabs}

\usepackage{mathtools}
\usepackage{amsthm}

\usepackage{multirow}

\usepackage[dvipsnames]{xcolor}

\theoremstyle{plain}

\theoremstyle{definition}

\theoremstyle{remark}

\definecolor{cvprblue}{rgb}{0.21,0.49,0.74}
\usepackage[pagebackref,breaklinks,colorlinks,citecolor=cvprblue]{hyperref}

\usepackage[accsupp]{axessibility}  

\usepackage{xcolor, soul}
\sethlcolor{pink}

\usepackage{pgfplots}
    \usepgfplotslibrary{colorbrewer}
    \pgfplotsset{
        cycle list/Dark2,
        cycle multiindex* list={
            mark list*\nextlist
            Dark2\nextlist
        },
    }
\pgfplotsset{compat=1.14}

\definecolor{ForestGreen}{RGB}{34,139,34}
\definecolor{Cerulean}{rgb}{0.0, 0.48, 0.65}
\definecolor{OrangeRed}{RGB}{255,69,0}
\definecolor{OrangeRed2}{RGB}{240,9,30}
\definecolor{OrangeRedSoft}{RGB}{240,40,50}
\definecolor{goldenpoppy}{rgb}{0.99, 0.76, 0.0}
\definecolor{goldenpoppy3}{rgb}{0.99,  0.46, 0.1}
\definecolor{skyblue}{rgb}{0.53, 0.81, 0.92}
\definecolor{skyblue2}{rgb}{0.55, 0.83, 0.94}
\definecolor{red-violet}{rgb}{0.78, 0.08, 0.52}
\definecolor{darkcerulean}{rgb}{0.03, 0.33, 0.55}
\definecolor{darkcerulean2}{rgb}{0.00, 0.20, 0.35}
\definecolor{flamingopink}{rgb}{0.99, 0.56, 0.67}
\definecolor{caribbeangreen}{rgb}{0.0, 0.8, 0.6}
\definecolor{caribbeangreen2}{rgb}{0.0, 0.825, 0.625}
\definecolor{darkpastelpurple}{rgb}{0.69, 0.34, 0.84}
\definecolor{darkpastelpurple2}{rgb}{0.68, 0.33, 0.83}
\definecolor{smokyblack}{rgb}{0.1, 0.1, 0.1}
\usepackage{standalone}

\usepackage{rotating}

\usepackage{booktabs}
\usepackage{array}
\newcolumntype{P}[1]{>{\raggedright\arraybackslash}p{#1}}
\usepackage{makecell}      

\usepackage{booktabs}
\usepackage{multirow}

\makeatletter
\renewcommand\paragraph{%
  \@startsection{paragraph}{4}{\z@}%
    {1.5mm}
    {-0.5em}
    {\normalfont\normalsize\bfseries}%
}
\makeatother

\usepackage{subcaption}
\title{SignalMC-MED: A Multimodal Benchmark for Evaluating Biosignal Foundation Models on Single-Lead ECG and PPG}

\author{Fredrik K. Gustafsson$^{1,2}$
\and
Xiao Gu$^{1}$
\and
Mattia Carletti$^{1}$
\and
Patitapaban Palo$^{1}$
\and
David W. Eyre$^{2,3,4}$
\and
David A. Clifton$^{1,2,5}$\vspace{1.0mm}
\and
\normalsize$^{1}$Department of Engineering Science, University of Oxford\vspace{-0.85mm}\\
\normalsize$^{2}$NIHR Health Protection Research Unit in Healthcare Associated Infections and Antimicrobial Resistance\vspace{-0.85mm}\\
\normalsize$^{3}$Big Data Institute, Nuffield Department of Population Health, University of Oxford\vspace{-0.85mm}\\
\normalsize$^{4}$NIHR Oxford Biomedical Research Centre\vspace{-0.85mm}\\
\normalsize$^{5}$Oxford Suzhou Centre for Advanced Research, University of Oxford\vspace{-0.6mm}\\
{\tt\small fredrik.gustafsson@eng.ox.ac.uk, xiao.gu@eng.ox.ac.uk, david.clifton@eng.ox.ac.uk}
}

\begin{document}

\maketitle

\begin{abstract}
    Recent biosignal foundation models (FMs) have demonstrated promising performance across diverse clinical prediction tasks, yet systematic evaluation on long-duration multimodal data remains limited. We introduce SignalMC-MED, a benchmark for evaluating biosignal FMs on synchronized single-lead electrocardiogram (ECG) and photoplethysmogram (PPG) data. Derived from the MC-MED dataset, SignalMC-MED comprises 22{,}256 visits with 10-minute overlapping ECG and PPG signals, and includes 20 clinically relevant tasks spanning prediction of demographics, emergency department disposition, laboratory value regression, and detection of prior ICD-10 diagnoses. Using this benchmark, we perform a systematic evaluation of representative time-series and biosignal FMs across ECG-only, PPG-only, and ECG + PPG settings. We find that domain-specific biosignal FMs consistently outperform general time-series models, and that multimodal ECG + PPG fusion yields robust improvements over unimodal inputs. Moreover, using the full 10-minute signal consistently outperforms shorter segments, and larger model variants do not reliably outperform smaller ones. Hand-crafted ECG domain features provide a strong baseline and offer complementary value when combined with learned FM representations. Together, these results establish SignalMC-MED as a standardized benchmark and provide practical guidance for evaluating and deploying biosignal FMs.
\end{abstract}
\vspace{-10.0mm}

\section*{}


\begin{figure*}[t]
    \centering
    \begin{subfigure}{0.9475\linewidth}
        \centering
        \includegraphics[width=\linewidth]{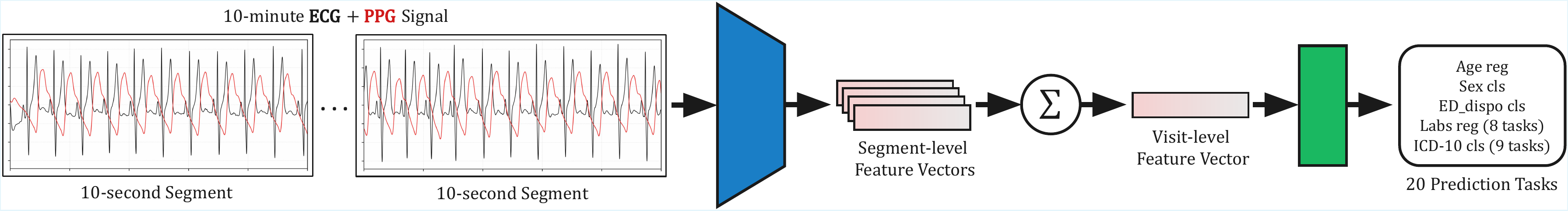}\vspace{1.0mm}
        \caption{\textbf{SignalMC-MED benchmark overview.}}
        \label{fig:overview}
    \end{subfigure}
    \vspace{0.4em}
    
    \begin{subfigure}{1.0\linewidth}
        \centering
        \includegraphics[width=\linewidth]{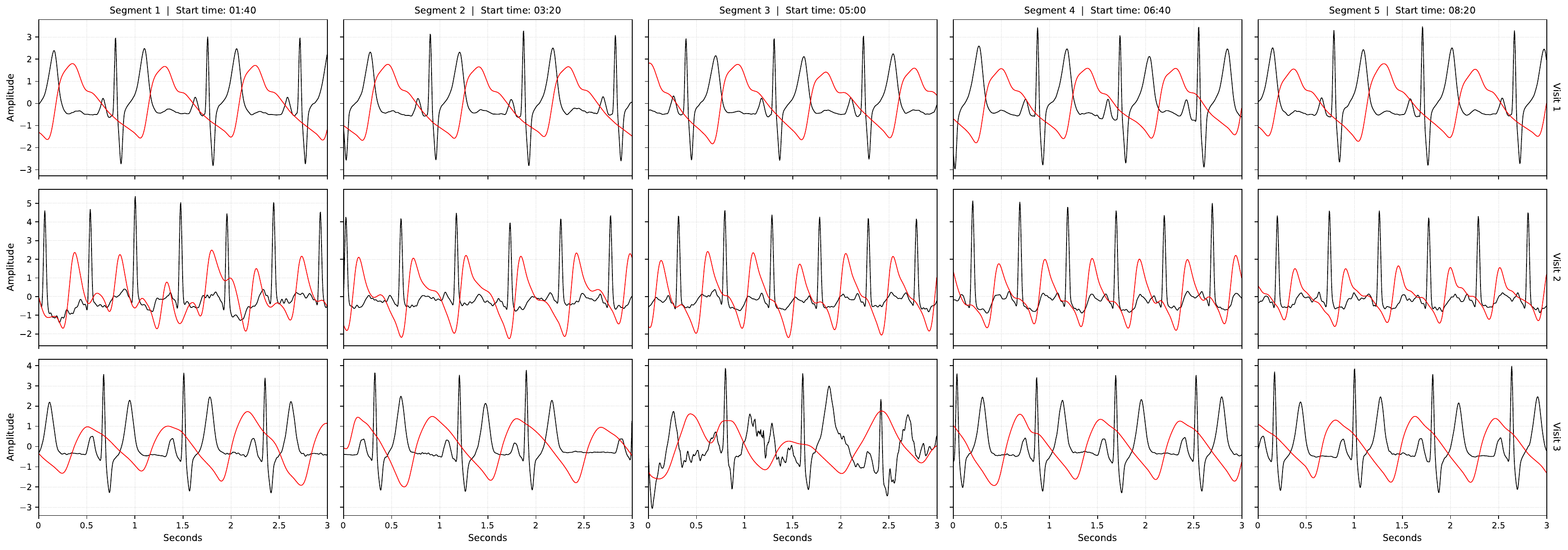}
        \caption{\textbf{Examples of synchronized ECG and PPG signals in SignalMC-MED.}}
        \label{fig:synced_ecg_ppg_signal_examples}
    \end{subfigure}\vspace{-1.0mm}
    
    \caption{\textbf{(a)} Overview of the SignalMC-MED benchmark and evaluation framework. For each of the 22{,}256 visits, synchronized 10-minute single-lead ECG (\textbf{black}) and PPG (\textcolor{OrangeRed2}{\textbf{red}}) signals are divided into non-overlapping 10-second segments. A frozen FM (\textcolor{Cerulean}{\textbf{blue}}) extracts a feature vector for each 10-second segment. The resulting segment-level features are aggregated via mean pooling ($\Sigma$) to form a visit-level representation, which is used to train a linear prediction model (\textcolor{ForestGreen}{\textbf{green}}) across 20 downstream tasks: age regression, sex classification, emergency department disposition classification, laboratory value regression (8 tasks), and prior ICD-10 diagnosis classification (9 tasks). \textbf{(b)} Examples of synchronized ECG and PPG signals from three different visits. For each visit, five 3-second segments are shown, evenly sampled from the full 10-minute signal (segment start times indicated above each panel). The signals illustrate inter-patient variability in rhythm, morphology, amplitude and noise characteristics. Additional examples are provided in Figure~\ref{fig:synced_ecg_ppg_signal_examples_more} in the supplementary material.}
    \label{fig:combined_overview}
\end{figure*}

Foundation models (FMs) for time-series and biosignals have recently emerged as a promising paradigm for scalable representation learning across diverse downstream tasks~\cite{bommasani2021opportunities, liang2024foundation, ye2024empowering, han2024systematic, gu2025foundation}. In cardiovascular medicine, large-scale pretraining on electrocardiogram (ECG) and photoplethysmography (PPG) signals offers the potential to learn physiologically meaningful representations that generalize across diagnostic and prognostic settings~\cite{li2025ecgfounder, saha2025pulse, abbaspourazad2024largescale, gu2025csfm}.

However, systematic evaluation on realistic, multimodal bedside data remains limited. In continuous monitoring environments, clinicians rely on long-duration single-lead ECG and PPG signals (often spanning minutes to hours in duration), which capture complementary and temporally evolving aspects of cardiovascular physiology~\cite{serhani2020ecg, bayoumy2021smart, sundrani2023predicting}. Yet, recent benchmarking studies~\cite{alcaraz2025enhancing, lunelli2025xecg, almasud2026benchmarking} and FM development papers~\cite{pillai2025papagei, saha2025pulse, thukral2026wavelet, hung2025d-beta, li2025ecgfounder, mckeen2025ecg} primarily focus on unimodal ECG or PPG data and short signal segments (typically 10 seconds). Moreover, while CSFM~\cite{gu2025csfm} demonstrates the promise of large-scale multimodal ECG + PPG pretraining across heterogeneous sensing scenarios, it does not establish a standardized benchmark explicitly designed for controlled comparison of FMs on synchronized long-duration ECG and PPG signals. Consequently, it remains unclear how different types of biosignal FMs perform on realistic multimodal emergency department data, and how these domain-specific FMs compare to general-purpose time-series FMs.  In particular, the interaction between multimodal fusion, varying signal lengths, and model scale has not been systematically evaluated, nor has the relative value of learned FM representations versus traditional domain feature baselines been rigorously assessed. 

To address these gaps, we introduce \textit{SignalMC-MED}, a clinically grounded benchmark derived from the MC-MED dataset~\cite{kansal2025mc, PhysioNet-mc-med-1.0.1}, comprising 22{,}256 adult emergency department visits with 10-minute synchronized ECG and PPG signals (Figure~\ref{fig:combined_overview}). The benchmark includes 20 clinically relevant tasks spanning prediction of demographics, emergency department disposition, laboratory value regression, and detection of \emph{prior} ICD-10 diagnoses, and adopts chronological train/val/test splits without patient overlap to approximate real-world deployment conditions. Using this benchmark, we perform a systematic evaluation of representative time-series and biosignal FMs (Table~\ref{table:fms}).

\textit{Our study provides three key contributions:}
(1) We introduce a large-scale multimodal benchmark for rigorous evaluation of biosignal FMs on synchronized ECG and PPG signals collected in routine emergency care.
(2) We conduct a systematic head-to-head evaluation of general time-series FMs, unimodal ECG- and PPG-specific biosignal FMs, a multimodal biosignal FM, and hand-crafted domain features across ECG-only, PPG-only, and ECG + PPG settings.
(3) We provide comprehensive empirical evidence clarifying how modality specialization, multimodal fusion, signal duration, model scale, and physiologically grounded domain features influence the performance of biosignal FMs.

\begin{table}[t]
	\caption{\textbf{Overview of all evaluated FMs}, including model size (number of parameters) and expected input sampling frequency.}\vspace{-2.0mm}	
    \label{table:fms}
    \centering
	\resizebox{1.0\linewidth}{!}{%
        \begin{tabular}{lcccP{4.75cm}}
\toprule
Model Name &Size &\makecell{Feature\\ Dimension} &\makecell{Sampling\\ Frequency} &Pretraining Data\\
\midrule
MOMENT-small~\citep{goswami2024moment} &35M  &512 &Mixed &\small{13M time series; 1.2B timestamps}\\
MOMENT-base                            &110M &768  &--- &---\\
MOMENT-large                           &341M &1024 &--- &---\\
\midrule
Chronos-Bolt-tiny~\citep{ansari2024chronos} &9M   &256 &Mixed &\small{100B time series observations}\\
Chronos-Bolt-mini                           &21M  &384 &--- &---\\
Chronos-Bolt-small                          &48M  &512 &--- &---\\
Chronos-Bolt-base                           &205M &768 &--- &---\\
\midrule
PaPaGei~\citep{pillai2025papagei} &6M &512 &125Hz &\small{\makecell[l]{20M 10-sec PPG segments\\ from 13.5K subjects}}\\
\midrule
D-BETA~\citep{hung2025d-beta} &62M &768 &500Hz &\small{\makecell[l]{0.78M 10-sec, 12-lead ECGs\\ from $\sim$150K subjects}}\\
\midrule
ECGFounder~\citep{li2025ecgfounder} &31M &1024 &500Hz &\small{\makecell[l]{10.8M ECGs (``predominantly 10-sec,\\ 12-lead'') from 1.8M subjects}}\\
\midrule
xECG~\citep{lunelli2025xecg} &57M &1024 &100Hz &\small{\makecell[l]{2.4M 7--10-sec, 12-lead ECGs\\ from $\sim$1.7M subjects\\ + 75 30-min, 12-lead ECGs}}\\
\midrule
CSFM-tiny~\citep{gu2025csfm} &43M  &768  &250Hz &\small{2.3M 10-sec, 12-lead ECGs + 0.27M}\\
CSFM-base                    &109M &768  &--- &\small{paired 10-sec PPGs and single-lead}\\
CSFM-large                   &334M &1024 &--- &\small{ECGs from 1.7M subjects}\\
\bottomrule
\end{tabular}




	}
\end{table}

\section*{Results}

\begin{table*}[t]
	\caption{\textbf{Main model comparison on the \textit{test} set}, with results aggregated across training using 10\%, 25\%, 50\%, and 100\% of the train visits. Regression tasks are evaluated using Pearson correlation ($\uparrow$), and classification tasks using AUROC ($\uparrow$). Columns correspond to aggregated performance for: age regression, sex classification, ED disposition classification, laboratory value regression (mean across eight laboratory tests), and prior ICD-10 diagnosis classification (mean across nine diagnosis groups). For each train percentage, training visits are sampled with replacement, and both downstream training and hyperparameter selection are repeated five times. Each table entry represents the mean $\pm$ standard deviation across resampling repetitions, after aggregation across tasks within each category and across train visit percentages. \textcolor{OrangeRed2}{\textbf{Bold red}} marks the best mean value in each column, \textbf{bold} marks the second-best, and \underline{underline} marks the third-best.}\vspace{-2.0mm}	
    \label{table:main_results_ecg-baseline-features-ppg-baseline-features-20sec_test}
    \centering
	\resizebox{1.0\linewidth}{!}{%
		\begin{tabular}{ll@{\hspace{0.8cm}}c@{\hspace{0.75cm}}c@{\hspace{0.75cm}}c@{\hspace{0.75cm}}c@{\hspace{0.75cm}}c}
\toprule
\textbf{Model Name} &\textbf{Modality} &\textbf{Age reg} &\textbf{Sex cls} &\textbf{ED\_dispo cls} &\textbf{Labs reg} &\textbf{ICD-10 cls}\\
\midrule
Domain Features &ECG-only                                       &0.678 $\pm$ 0.002    &0.652 $\pm$ 0.004    &0.711 $\pm$ 0.003    &0.319 $\pm$ 0.002    &\textcolor{OrangeRed2}{\textbf{0.748}} $\pm$ 0.003\\
Domain Features &PPG-only                                       &0.684 $\pm$ 0.052    &0.737 $\pm$ 0.002    &0.676 $\pm$ 0.002    &0.213 $\pm$ 0.018    &0.677 $\pm$ 0.002\\
Domain Features (concat)\hspace{4.0mm} &ECG + PPG\hspace{10.0mm} &0.751 $\pm$ 0.044    &0.769 $\pm$ 0.002    &\textbf{0.713} $\pm$ 0.001    &0.299 $\pm$ 0.015    &\textbf{0.736} $\pm$ 0.005\\
\midrule
MOMENT-base &ECG-only  &0.641 $\pm$ 0.002    &0.741 $\pm$ 0.002    &0.662 $\pm$ 0.004    &0.239 $\pm$ 0.005    &0.671 $\pm$ 0.005\\
MOMENT-base &PPG-only  &0.702 $\pm$ 0.002    &0.724 $\pm$ 0.003    &0.676 $\pm$ 0.002    &0.240 $\pm$ 0.002    &0.670 $\pm$ 0.002\\
MOMENT-base &ECG + PPG &0.757 $\pm$ 0.002    &0.776 $\pm$ 0.001    &0.678 $\pm$ 0.003    &0.283 $\pm$ 0.004    &0.692 $\pm$ 0.004\\
\midrule
Chronos-Bolt-small &ECG-only  &0.650 $\pm$ 0.004    &0.720 $\pm$ 0.002    &0.652 $\pm$ 0.004    &0.233 $\pm$ 0.002    &0.666 $\pm$ 0.004\\
\bottomrule
\end{tabular}
	}
\end{table*}

We evaluate eight representative approaches spanning two general time-series FMs (MOMENT~\citep{goswami2024moment}, Chronos-Bolt~\citep{ansari2024chronos}), four unimodal biosignal FMs (D-BETA~\citep{hung2025d-beta}, ECGFounder~\citep{li2025ecgfounder}, xECG~\citep{lunelli2025xecg}, PaPaGei~\citep{pillai2025papagei}), one multimodal biosignal FM (CSFM~\citep{gu2025csfm}), and hand-crafted ECG and PPG domain features. None of the evaluated FMs were pretrained on MC-MED, preventing any overlap between pretraining data and the benchmark. All FMs are evaluated in ECG-only, PPG-only, and ECG + PPG settings, including those pretrained exclusively on a single modality. This unified evaluation allows us to assess cross-modality generalization and determine whether learned representations encode modality-specific morphology or more general cardiovascular dynamics transferable across related biosignals. 

All models are used strictly as frozen feature extractors within a unified linear-probing framework to isolate representation quality (see \textit{Methods} and Figure~\ref{fig:method}). Each 10-minute signal is divided into non-overlapping 10-second segments, segment-level features are extracted, and visit-level representations are obtained via mean aggregation. For shorter-duration experiments (10 seconds to 5 minutes), only the initial contiguous portion of the signal is used. Downstream prediction is performed using ridge regression for regression tasks and regularized logistic regression for classification tasks. For each model, a single regularization hyperparameter per task type is selected on the \textit{val} split.

Performance is reported using Pearson correlation for regression and area under the receiver operating characteristic curve (AUROC) for classification. To assess robustness and data efficiency, downstream models are trained on 10\%, 25\%, 50\%, and 100\% of the available \textit{train} visits with five repeated resamplings. Multimodal fusion is implemented via late feature-level fusion by averaging independently extracted ECG-only and PPG-only visit-level representations. The 20 benchmark tasks (Table~\ref{table:csfm-base_results_all-20-tasks_val}) are grouped into five categories and models are ranked by mean category rank to facilitate cross-model comparison.

\subsubsection*{Main Model Comparison}

Table~\ref{table:main_results_ecg-baseline-features-ppg-baseline-features-20sec_all-models_val} in the supplementary material reports \textit{val} results for all model variants, with corresponding rankings in Table~\ref{table:main_results_ecg-baseline-features-ppg-baseline-features-20sec_all-models_val_rank}. Based on these, one representative variant each for MOMENT, Chronos-Bolt, xECG, CSFM, and the PPG domain features is selected and evaluated on the \textit{test} set alongside D-BETA, PaPaGei, and ECG domain features (Table~\ref{table:main_results_ecg-baseline-features-ppg-baseline-features-20sec_test}). Model rankings on \textit{test} are shown separately for ECG-only, PPG-only, and ECG + PPG in Table~\ref{table:main_results_ecg-baseline-features-ppg-baseline-features-20sec_test_rank_subsets}, and jointly across modalities in Table~\ref{table:main_results_ecg-baseline-features-ppg-baseline-features-20sec_test_rank}. Moreover, Figure~\ref{figure:main_results_ecg-baseline-features-ppg-baseline-features-20sec_test} shows the \textit{test} performance as a function of the percentage of train visits used, highlighting both absolute performance and data-efficiency trends.

Across modalities, CSFM-base achieves the best overall performance and mean rank, while xECG consistently attains the second-best performance. The two models perform strongly across both regression and classification tasks. In contrast, PaPaGei and D-BETA rank among the weakest performers overall. Notably, PaPaGei, which is the only evaluated PPG-specific FM, ranks last for ECG-only and ECG + PPG, and second-last even for PPG-only. Hand-crafted domain features provide a strong baseline, ranking fourth across all modalities and outperforming multiple FMs. Among ECG-specific models, ECGFounder ranks third for ECG-only and ECG + PPG but drops for PPG-only, whereas xECG, despite ECG-only pretraining, ranks second across modalities and performs strongly even on PPG-only inputs.

Table~\ref{table:csfm-base_results_all-20-tasks_val} further reports CSFM-base \textit{val} performance across all 20 benchmark tasks. Performance varies substantially across tasks, with particularly strong results for atrial fibrillation and cardiac device detection, and more modest performance for conditions such as anemia and venous thromboembolism. Results on \textit{val} for all eight evaluated models across the 20 tasks, when training with 10\%, 25\%, 50\%, and 100\% of the train visits, are provided in Table~\ref{table:results_all-20-tasks_val_ecg_std}~-~\ref{table:results_all-20-tasks_val_ecg_ppg_mean_std_10}. Across tasks and train visit percentages, variability across resampling repetitions remains small overall, indicating stable performance estimates and consistent model rankings.

\subsubsection*{Extended Model Analysis}

Beyond the primary benchmark comparison, we further analyze model behavior under different modality settings, signal lengths, and architectural configurations. Figure~\ref{figure:results_ecg-baseline-features-ppg-baseline-features-20sec_val_fusion}~\&~\ref{figure:results_ecg-baseline-features-ppg-baseline-features-20sec_val_fusion_part2} show that ECG + PPG fusion consistently improves performance over unimodal inputs for nearly all models and train percentages. Figure~\ref{figure:results_ecg-baseline-features-ppg-baseline-features-20sec_val_fusion-signal-lens}~\&~\ref{figure:results_ecg-baseline-features-ppg-baseline-features-20sec_val_fusion-signal-lens_part2} demonstrate that these gains persist across different signal lengths, indicating that multimodal complementarity is not limited to specific temporal resolutions. 

\begin{figure}[t]
    \centering
    \begin{minipage}{1.0\linewidth}
        \centering
        \captionof{table}{\textbf{Main model ranking on the \textit{test} set}, computed separately for (a) ECG-only, (b) PPG-only, and (c) ECG + PPG inputs. Mean model rank ($\downarrow$) across the five aggregated task categories based on Table~\ref{table:main_results_ecg-baseline-features-ppg-baseline-features-20sec_test}. A joint ranking across modalities is in Table~\ref{table:main_results_ecg-baseline-features-ppg-baseline-features-20sec_test_rank}.}\vspace{-1.0mm}
        \label{table:main_results_ecg-baseline-features-ppg-baseline-features-20sec_test_rank_subsets}
        \begin{subtable}{1.0\linewidth}
        \centering
        \vskip -0.05in
        \resizebox{1.0\linewidth}{!}{
        \begin{tabular}{clcl}
\toprule
\textbf{Rank} &\textbf{Model Name} &\textbf{Model Ranks} &\textbf{Mean Model Rank ($\downarrow$)}\\
\midrule
1    &CSFM-base        &7,5,5,2,4        &\hspace{13.0mm}4.6\\
2    &xECG-10min        &10,3,7,4,8        &\hspace{13.0mm}6.4\\
3    &ECGFounder        &11,12,9,7,7        &\hspace{13.0mm}9.2\\
4    &Domain Features        &17,23,4,6,1        &\hspace{13.0mm}10.2\\
5    &MOMENT-base        &21,11,19,15,16        &\hspace{13.0mm}16.4\\
6    &D-BETA        &22,21,18,20,12        &\hspace{13.0mm}18.6\\
7    &Chronos-Bolt-small\hspace{6.0mm}        &19,18,20,18,19        &\hspace{13.0mm}18.8\\
8    &PaPaGei        &24,24,24,23,24        &\hspace{13.0mm}23.8\\
\bottomrule
\end{tabular}
        }\vspace{0.25mm}
        \caption{ECG-only.}
        \label{table:main_results_ecg-baseline-features-ppg-baseline-features-20sec_test_rank_ecg}
        \end{subtable}
        \vskip 0.05in
        \begin{subtable}{1.0\linewidth}
        \centering
        \vskip -0.05in
        \resizebox{1.0\linewidth}{!}{
        \begin{tabular}{clcl}
\toprule
\textbf{Rank} &\textbf{Model Name} &\textbf{Model Ranks} &\textbf{Mean Model Rank ($\downarrow$)}\\
\midrule
1    &CSFM-base        &4,9,3,8,9        &\hspace{13.0mm}6.6\\
2    &xECG-10min        &6,10,10,12,14        &\hspace{13.0mm}10.4\\
3    &MOMENT-base        &12,17,12,13,17        &\hspace{13.0mm}14.2\\
4    &Domain Features        &15,13,13,22,15        &\hspace{13.0mm}15.6\\
5    &Chronos-Bolt-small\hspace{6.0mm}        &13,15,17,16,22        &\hspace{13.0mm}16.6\\
6    &ECGFounder        &16,16,14,19,18        &\hspace{13.0mm}16.6\\
7    &PaPaGei        &18,22,22,17,21        &\hspace{13.0mm}20.0\\
8    &D-BETA        &23,20,21,24,23        &\hspace{13.0mm}22.2\\
\bottomrule
\end{tabular}
        }\vspace{0.25mm}
        \caption{PPG-only.}
        \label{table:main_results_ecg-baseline-features-ppg-baseline-features-20sec_test_rank_ppg}
        \end{subtable}
        \vskip 0.05in
        \begin{subtable}{1.0\linewidth}
        \centering
        \vskip -0.05in
        \resizebox{1.0\linewidth}{!}{
        \begin{tabular}{clcl}
\toprule
\textbf{Rank} &\textbf{Model Name} &\textbf{Model Ranks} &\textbf{Mean Model Rank ($\downarrow$)}\\
\midrule
1    &CSFM-base        &1,1,1,1,3        &\hspace{13.0mm}1.4\\
2    &xECG-10min        &2,2,6,3,6        &\hspace{13.0mm}3.8\\
3    &ECGFounder        &3,4,8,5,5        &\hspace{13.0mm}5.0\\
4    &Domain Features (concat)\hspace{6.0mm}        &8,7,2,9,2        &\hspace{13.0mm}5.6\\
5    &MOMENT-base        &5,6,11,10,11        &\hspace{13.0mm}8.6\\
6    &Chronos-Bolt-small        &9,8,16,11,13        &\hspace{13.0mm}11.4\\
7    &D-BETA        &14,14,15,14,10        &\hspace{13.0mm}13.4\\
8    &PaPaGei        &20,19,23,21,20        &\hspace{13.0mm}20.6\\
\bottomrule
\end{tabular}
        }\vspace{0.25mm}
        \caption{ECG + PPG.}
        \label{table:main_results_ecg-baseline-features-ppg-baseline-features-20sec_test_rank_ecg_ppg_mean}
        \end{subtable}
    \end{minipage}
\end{figure}

Using the full 10-minute signal consistently outperforms shorter segments (Figure~\ref{figure:results_ecg-baseline-features-ppg-baseline-features-20sec_val_signal-lens},~\ref{figure:results_ecg-baseline-features-ppg-baseline-features-20sec_val_signal-lens_part2}~\&~\ref{figure:results_ecg-baseline-features-ppg-baseline-features-20sec_val_signal-lens_domain-features}), suggesting that longer temporal context provides additional predictive information beyond short-window morphology. In contrast, larger variants within the same model family (e.g., CSFM-tiny/base/large) do \textit{not} reliably outperform their smaller counterparts (Figure~\ref{figure:results_ecg-baseline-features-ppg-baseline-features-20sec_val_model-sizes}~\&~\ref{figure:results_ecg-baseline-features-ppg-baseline-features-20sec_val_model-sizes_pca-256}), indicating diminishing returns from parameter scaling under frozen linear evaluation.

Figure~\ref{figure:results_ecg-baseline-features-ppg-baseline-features-20sec_val_domain-features-concat_comp}~\&~\ref{figure:results_ecg-baseline-features-ppg-baseline-features-20sec_val_domain-features-concat_comp_ppg} evaluate concatenation of hand-crafted domain features with learned FM representations. For ECG, domain-feature augmentation consistently improves performance, whereas improvements are less consistent for PPG. Moreover, simple feature-level fusion slightly outperforms CSFM's internal multimodal mechanism (Figure~\ref{figure:results_ecg-baseline-features-ppg-baseline-features-20sec_val_csfm-fusion}), and the performance gap between CSFM-base and domain features narrows with longer signals (Figure~\ref{figure:results_ecg-baseline-features-ppg-baseline-features-20sec_val_csfm-base_domain-features_comp}), suggesting that temporal aggregation partially compensates for representational differences.

\begin{figure*}[t]
\centering
    \begin{subfigure}[t]{0.33\textwidth}
        \centering%
        \includegraphics[clip, trim=1.5cm 0.95cm 1.85cm 0.95cm, width=1.0\linewidth]{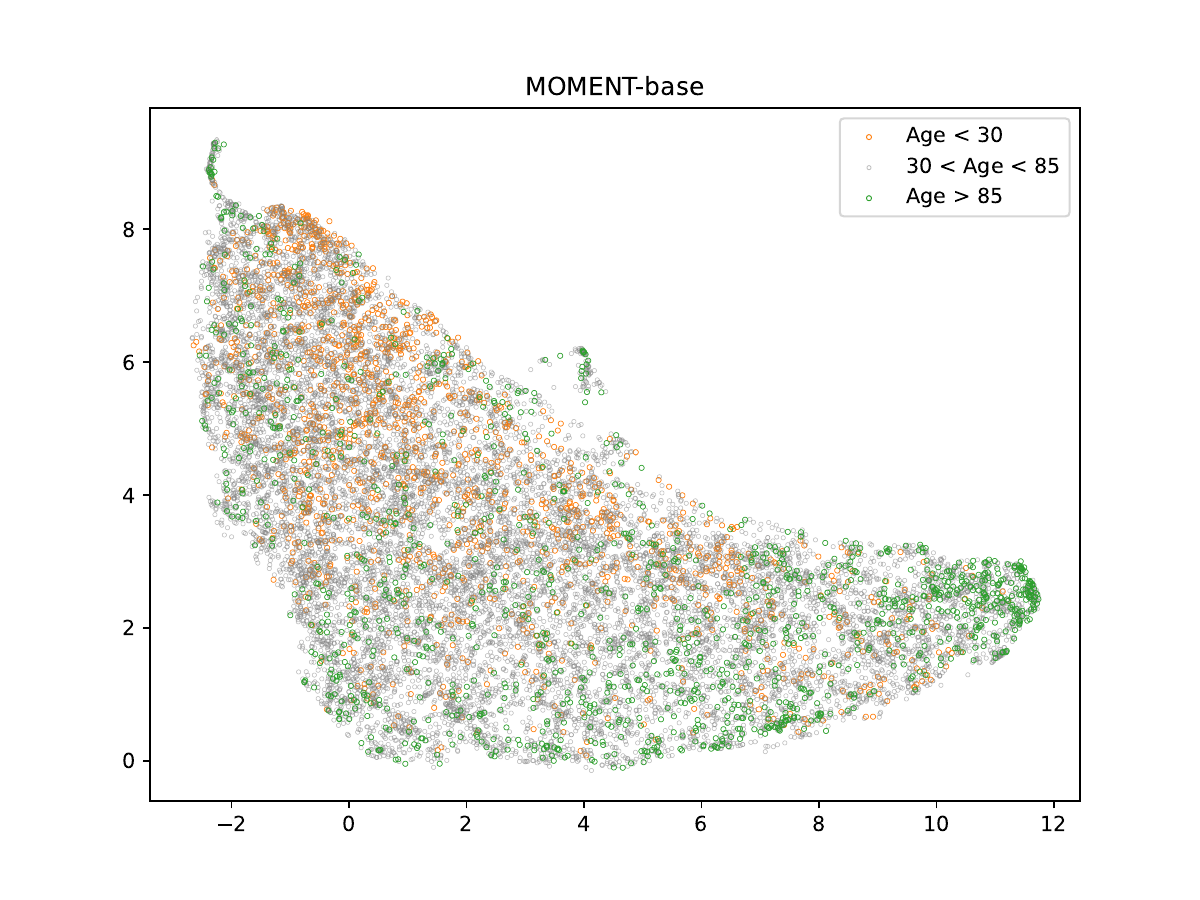}
    \end{subfigure}
    \begin{subfigure}[t]{0.33\textwidth}
        \centering%
        \includegraphics[clip, trim=1.5cm 0.95cm 1.85cm 0.95cm, width=1.0\linewidth]{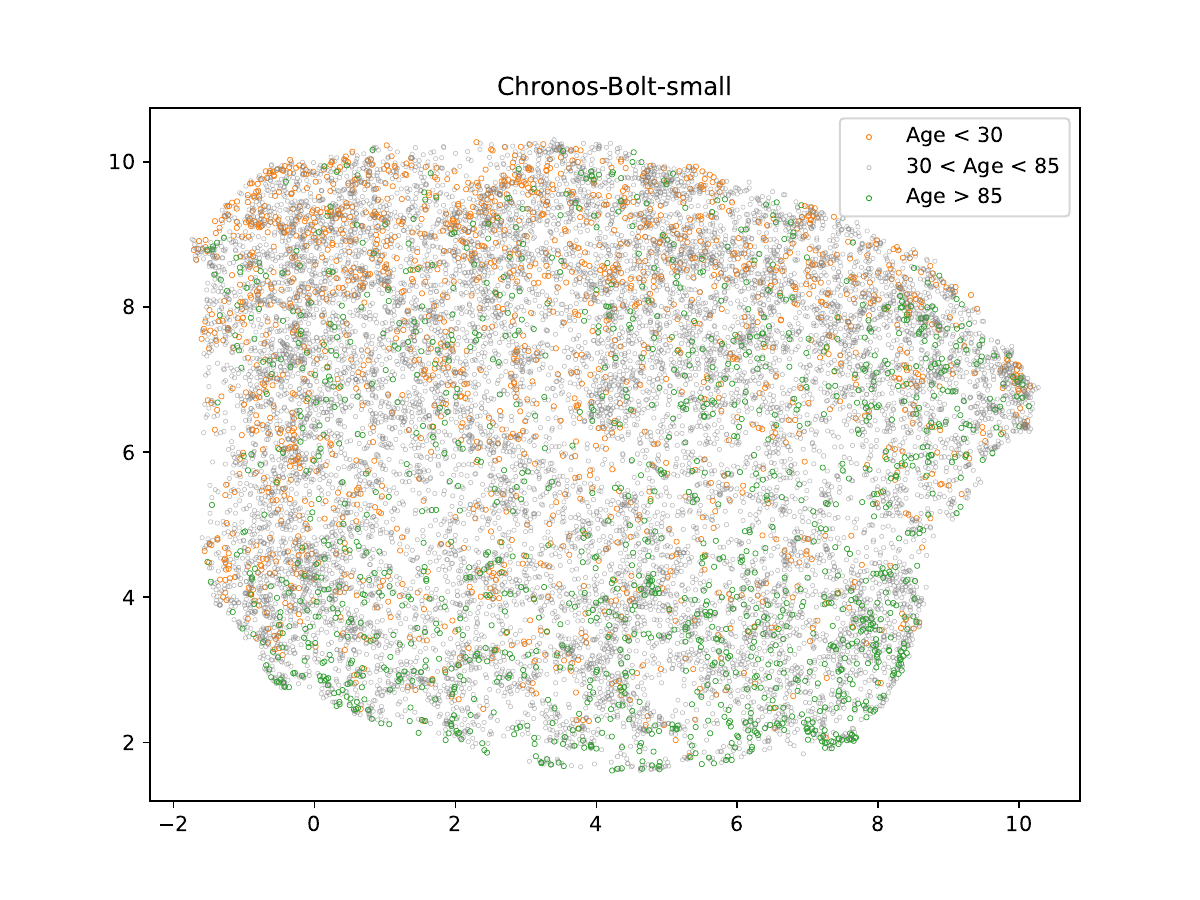}
    \end{subfigure}
    \begin{subfigure}[t]{0.33\textwidth}
        \centering%
        \includegraphics[clip, trim=1.5cm 0.95cm 1.85cm 0.95cm, width=1.0\linewidth]{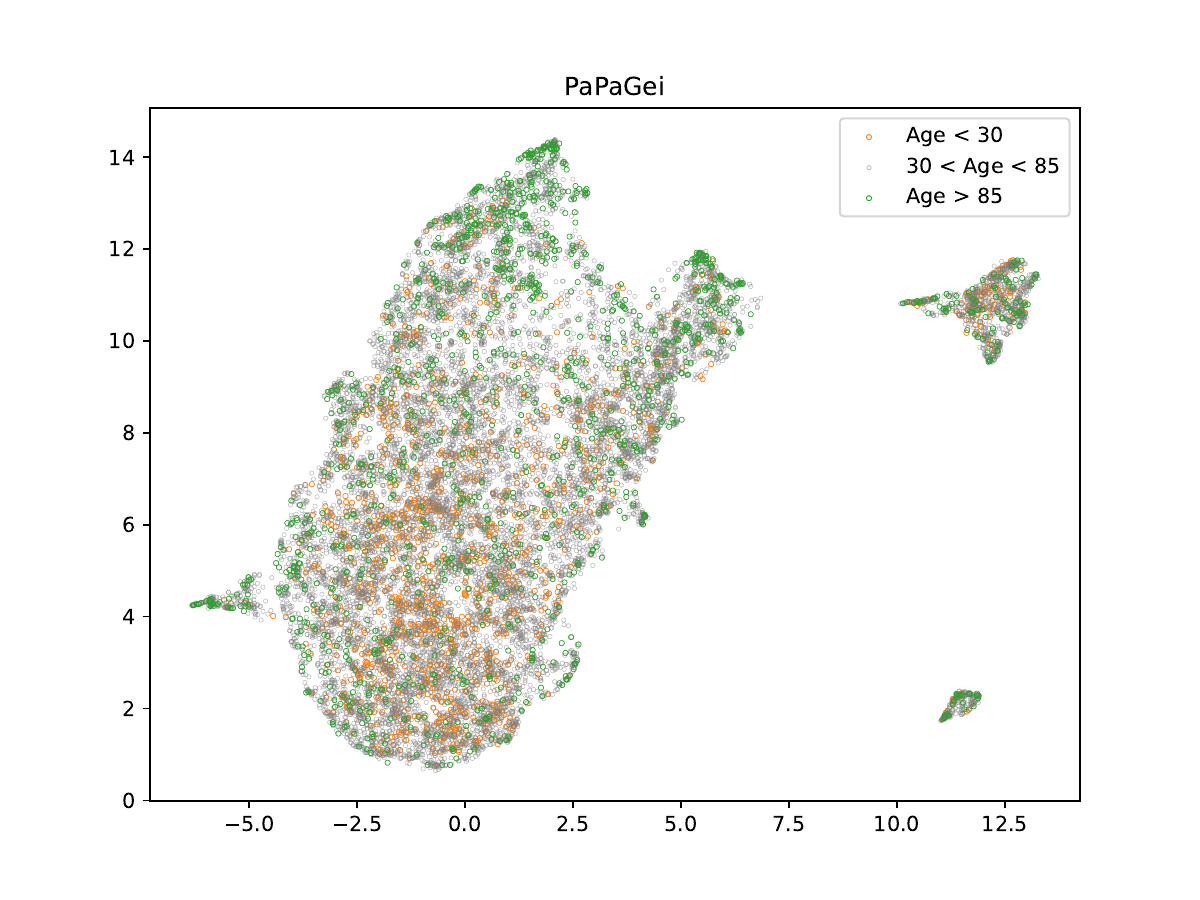}
    \end{subfigure}
    \begin{subfigure}[t]{0.33\textwidth}
        \centering%
        \includegraphics[clip, trim=1.5cm 0.95cm 1.85cm 0.95cm, width=1.0\linewidth]{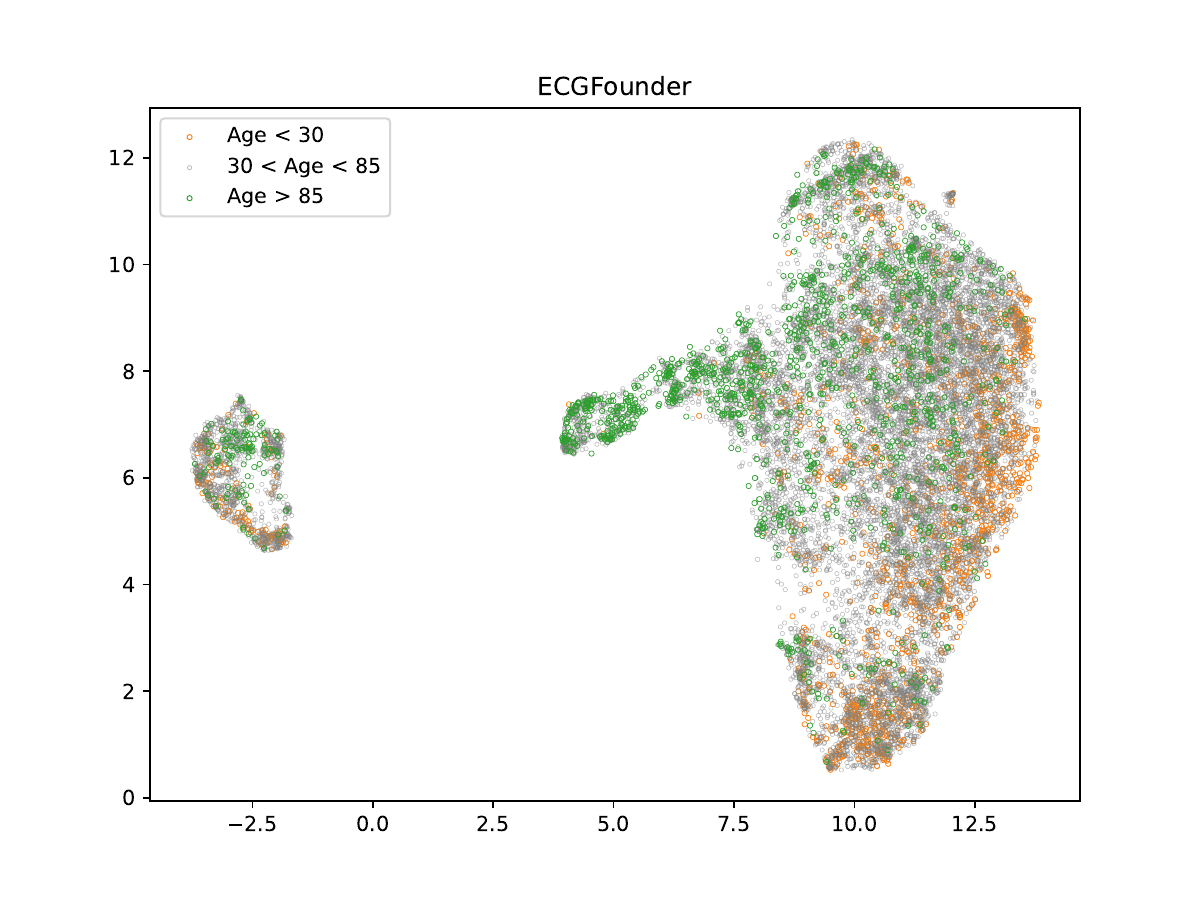}
    \end{subfigure}
    \begin{subfigure}[t]{0.33\textwidth}
        \centering%
        \includegraphics[clip, trim=1.5cm 0.95cm 1.85cm 0.95cm, width=1.0\linewidth]{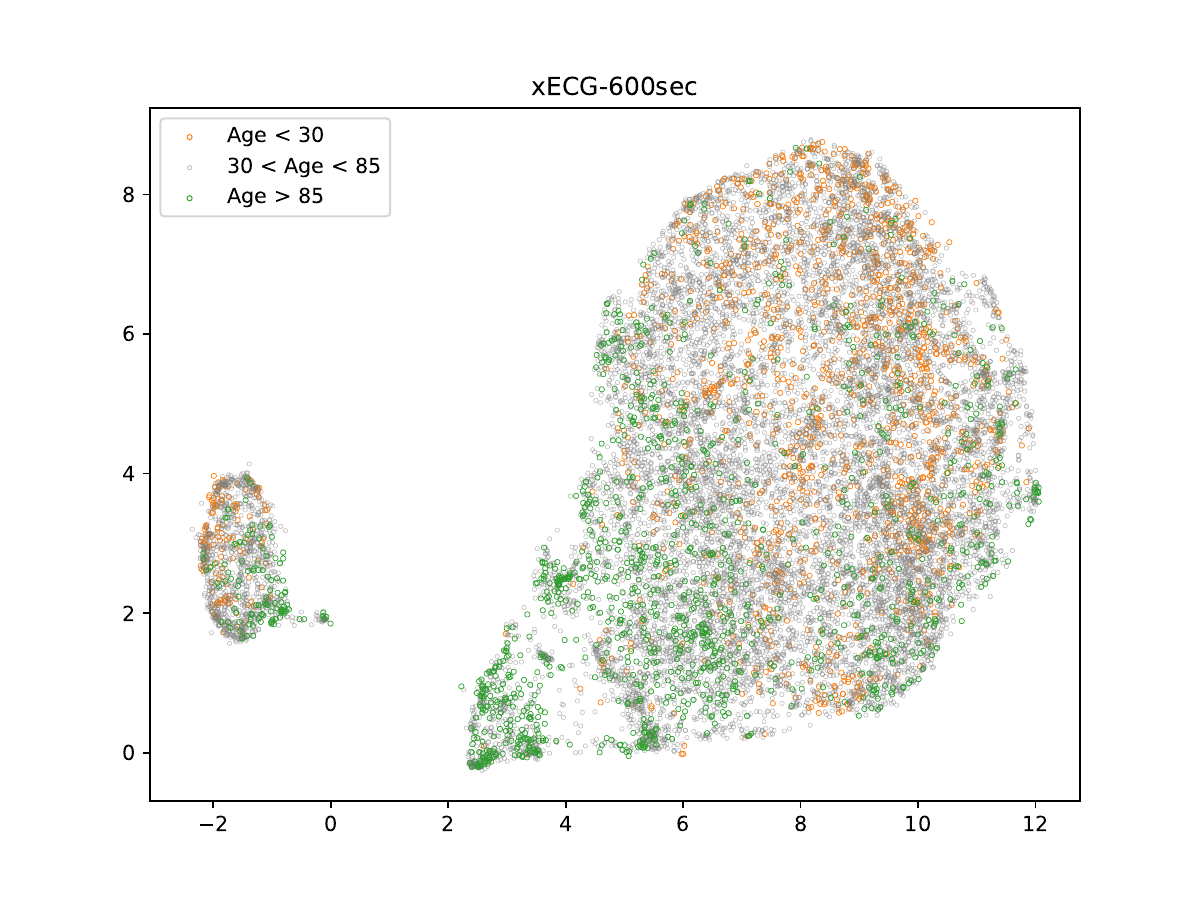}
    \end{subfigure}
    \begin{subfigure}[t]{0.33\textwidth}
        \centering%
        \includegraphics[clip, trim=1.5cm 0.95cm 1.85cm 0.95cm, width=1.0\linewidth]{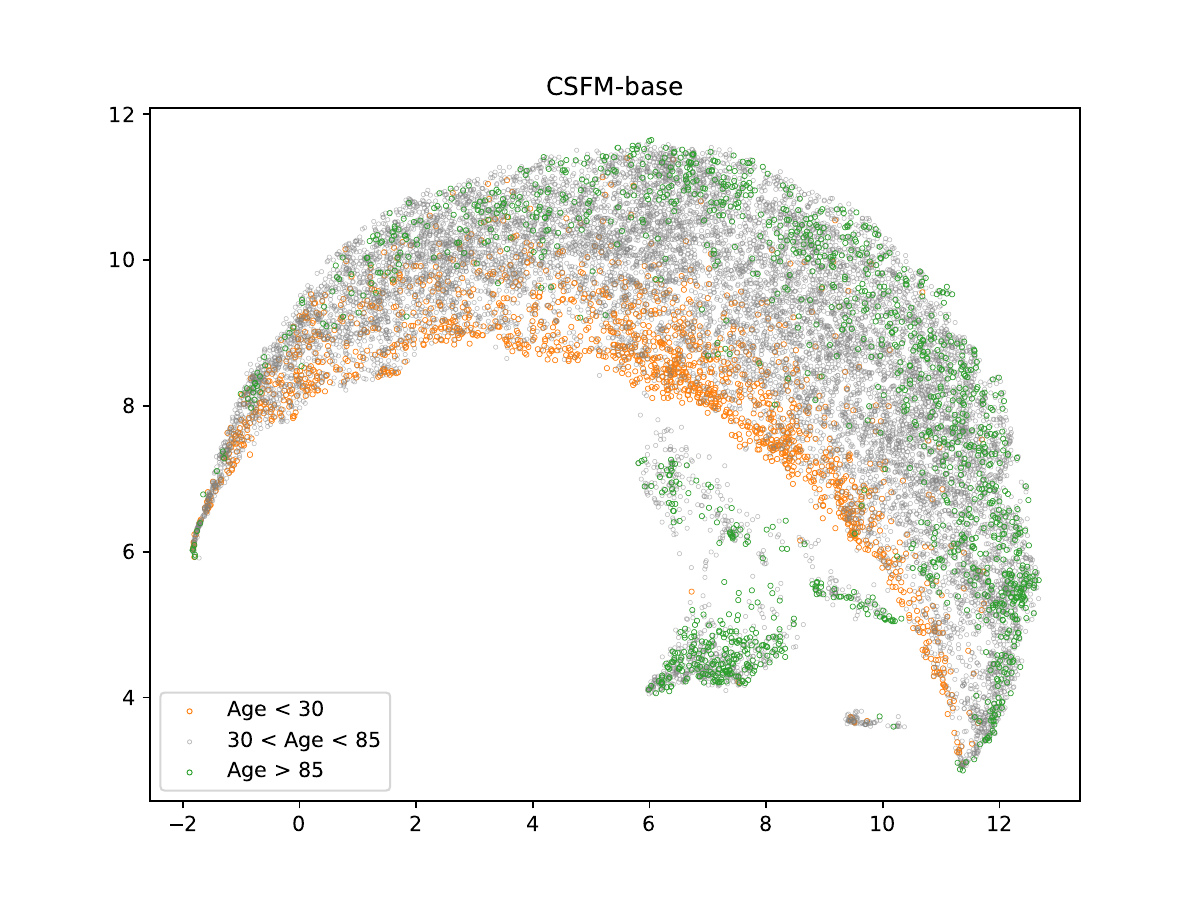}
    \end{subfigure}
  \caption{\textbf{UMAP visualizations colored by age group}. UMAP~\citep{mcinnes2018umap} projections of the \textit{train} set \textit{ECG + PPG} feature vectors for six selected models, colored according to three age groups ($<30$ years, $30-85$ years, $>85$ years). CSFM-base, xECG-10min and ECGFounder exhibit clearer age-related structure compared to general time-series FMs (MOMENT-base, Chronos-Bolt-small) and PaPaGei.}
  \label{fig:umap_plots_age_groups}
\end{figure*}

Finally, UMAP visualizations in Figure~\ref{fig:umap_plots_age_groups}~\&~\ref{fig:umap_plots_I48} show clearer clustering by age group and atrial fibrillation label for CSFM-base, xECG, and ECGFounder than for general time-series FMs, consistent with their stronger quantitative performance. Although UMAP provides only a qualitative projection, these patterns suggest that domain-specific biosignal FMs can learn more structured and clinically aligned latent representations.

\begin{table*}[t]
	\caption{\textbf{SignalMC-MED benchmark task summary and CSFM-base reference performance}. Overview of the 20 included tasks, grouped into two demographics tasks, ED disposition classification, eight laboratory value regression tasks, and nine prior ICD-10 diagnosis classification tasks. CSFM-base performance across modalities on the \textit{val} set when using \textit{100\%} of the train set visits is shown as a reference for relative task difficulty. Training visits are sampled with replacement, and both downstream training and hyperparameter selection are repeated five times; results are reported as mean $\pm$ standard deviation across resampling repetitions. \textcolor{OrangeRed2}{\textbf{Bold red}} marks the best mean value in each row. Full results with 100\%, 50\%, 25\%, and 10\% of the train visits for all eight evaluated models are provided in Table~\ref{table:results_all-20-tasks_val_ecg_std} - \ref{table:results_all-20-tasks_val_ecg_ppg_mean_std_10}.}\vspace{-2.0mm}	
    \label{table:csfm-base_results_all-20-tasks_val}
    \centering
	\resizebox{1.0\linewidth}{!}{%
        \begin{tabular}{lllccc}
\toprule
\textbf{Task} &\textbf{Label / Description} &\textbf{Label Source} &\textbf{ECG-only} &\textbf{PPG-only} &\textbf{ECG + PPG} \\
\midrule
Age reg &``Age'' &\multirow{2}{*}{\texttt{visits.csv}} &0.785 $\pm$ 0.002    &0.799 $\pm$ 0.002    &\textcolor{OrangeRed2}{\textbf{0.842}} $\pm$ 0.001\\
Sex cls &``M'' vs ``F'' &                              &0.816 $\pm$ 0.002    &0.767 $\pm$ 0.003    &\textcolor{OrangeRed2}{\textbf{0.840}} $\pm$ 0.002\\
\midrule
ED\_dispo cls &``Discharge'' vs \{``Inpatient'', ``Observation'', ``ICU''\}\hspace{5.0mm} &\texttt{visits.csv}\hspace{10.0mm} &0.744 $\pm$ 0.001    &0.726 $\pm$ 0.001    &\textcolor{OrangeRed2}{\textbf{0.750}} $\pm$ 0.002\\
\midrule
Potassium reg &``POTASSIUM'' &\multirow{8}{*}{\texttt{labs.csv}} &\textcolor{OrangeRed2}{\textbf{0.386}} $\pm$ 0.007    &0.190 $\pm$ 0.006    &\textcolor{OrangeRed2}{\textbf{0.386}} $\pm$ 0.006\\
Calcium reg &``CALCIUM''&                                        &0.209 $\pm$ 0.010    &0.183 $\pm$ 0.015    &\textcolor{OrangeRed2}{\textbf{0.265}} $\pm$ 0.010\\
eGFR reg &``EGFR REFIT WITHOUT RACE (2021)''&                    &0.591 $\pm$ 0.006    &0.567 $\pm$ 0.005    &\textcolor{OrangeRed2}{\textbf{0.613}} $\pm$ 0.003\\
Glucose reg & `GLUCOSE''&                                        &0.262 $\pm$ 0.017    &0.297 $\pm$ 0.009    &\textcolor{OrangeRed2}{\textbf{0.322}} $\pm$ 0.010\\
\bottomrule
\end{tabular}
	}
\end{table*}

\input{figures/main_results_ecg-baseline-features-ppg-baseline-features-20sec_test}

\input{figures/results_ecg-baseline-features-ppg-baseline-features-20sec_val_fusion}

\input{figures/results_ecg-baseline-features-ppg-baseline-features-20sec_val_signal-lens}

\input{figures/results_ecg-baseline-features-ppg-baseline-features-20sec_val_fusion-signal-lens}

\input{figures/results_ecg-baseline-features-ppg-baseline-features-20sec_val_domain-features-concat_comp}

\section*{Discussion}

Overall, our results demonstrate a clear advantage of domain-specific biosignal FMs over general time-series FMs across ECG-only, PPG-only, and ECG + PPG settings. CSFM-base achieves the best overall performance with xECG consistently second. This suggests that while general time-series FMs may learn broadly transferable sequence patterns, they lack inductive biases tailored to cardiovascular physiology. However, the poor performance of PaPaGei even in the PPG-only setting shows that modality-specific pretraining alone does not guarantee strong performance. Notably, although PaPaGei is pretrained on 20 million PPG segments, these originate from only 13{,}500 subjects, whereas CSFM, ECGFounder, and xECG are trained on data from more than 1.5 million individuals (Table~\ref{table:fms}). The substantially greater subject-level diversity in the latter models may expose them to a wider range of physiological variability and signal conditions, which likely contributes to improved downstream generalization beyond what can be achieved through segment-level scale alone.

Multimodal ECG + PPG fusion yields robust and consistent gains across tasks, train percentages, and signal lengths, including for models pretrained exclusively on ECG, highlighting that PPG provides complementary physiological information beyond what can be captured from ECG alone. Notably, the simple late-fusion strategy of averaging unimodal features slightly outperforms CSFM's internal multimodal mechanism. This may reflect several factors, including the relative scarcity of paired ECG + PPG segments compared to ECG-only data in CSFM's multimodal pretraining (Table~\ref{table:fms}), and the absence of explicit cross-modality alignment losses. Under frozen linear evaluation, preserving unimodal structure and combining it at the feature level may also allow the downstream model to exploit complementary information more flexibly than relying on a fixed internal fusion mechanism. These findings suggest that effective multimodal integration remains a non-trivial design challenge.

Signal duration proves more impactful than architectural scale under our frozen linear evaluation: extending inputs to the full 10-minute recordings produces larger and more reliable improvements than increasing parameter count, and larger model variants do not consistently outperform mid-sized models. When representations are projected to a 256-dimensional space using PCA, performance trends remain similar (Figure~\ref{figure:results_ecg-baseline-features-ppg-baseline-features-20sec_val_model-sizes}~\&~\ref{figure:results_ecg-baseline-features-ppg-baseline-features-20sec_val_model-sizes_pca-256}), indicating that the occasional underperformance of larger model variants is not attributable to differences in feature dimensionality alone. This scaling saturation may instead reflect a representational bottleneck under linear probing: once key rhythm and morphology features are linearly accessible, additional capacity may encode information that is not exploitable without fine-tuning. It may also indicate that model size is not the primary constraint at current pretraining data scales.

Hand-crafted domain features provide a strong baseline and usually outperform both MOMENT and Chronos-Bolt, highlighting the continued value of physiologically grounded feature engineering. Importantly, feature concatenation experiments further reveal complementary strengths between learned and hand-crafted representations. For ECG, augmenting FM embeddings with domain features consistently improves performance, even for models that already outperform standalone domain features, suggesting that explicit morphological descriptors add information beyond learned FM representations. The comparatively weaker and less consistent gains for PPG indicate that our ECG-derived features encode particularly informative structure. These findings suggest that current ECG FMs may not fully internalize certain established morphological abstractions, and that incorporating auxiliary objectives or physiologically informed inductive biases during pretraining could help ensure that learned FM representations subsume strong handcrafted features rather than merely complement them.

Interestingly, xECG performs strongly even on PPG-only inputs despite ECG-only pretraining, suggesting partial transferability of learned temporal and rhythm-level representations across related cardiovascular signals. Both ECG and PPG reflect manifestations of the same underlying cardiac cycle, and models trained to capture long-range temporal structure and inter-beat dynamics in ECG may therefore learn higher-level cardiovascular representations that remain informative for PPG. In contrast, the PPG-specific model PaPaGei does not generalize well beyond its training modality, which may in part reflect the limited subject-level diversity in its pretraining data. Together, these findings suggest that large-scale exposure to diverse cardiovascular dynamics may be more important for cross-modality robustness than strict modality specialization alone.

Finally, variability in classification task performance does not follow class prevalence alone (Table~\ref{table:cls_tasks}). For example, cardiac device detection (2.4\% \textit{test} prevalence) and atrial fibrillation detection (7.6\%) achieve among the highest AUROC values, whereas more prevalent conditions such as anemia (11.1\%) and respiratory compromise (13.1\%) show more modest performance. This indicates that predictive accuracy is driven primarily by the strength and specificity of physiological signatures present in ECG and PPG signals rather than by label frequency. Indeed, conditions such as atrial fibrillation and cardiac device implantation are known to produce clear and direct waveform-level manifestations, whereas anemia does not have a specific electrocardiographic signature. Venous thromboembolism may also occur without overt cardiac strain, consistent with the more modest performance observed for this task.

This study has several limitations. All evaluations rely on frozen representations with linear downstream models, and results may differ under end-to-end fine-tuning or nonlinear adaptation. The benchmark focuses on the first 10 minutes of ECG and PPG recordings and employs simple temporal aggregation strategies. Future work should extend the evaluation to continuous monitoring scenarios with hours or days of data per patient to better assess long-term temporal modeling. We evaluate seven representative FMs spanning general time-series models and both unimodal and multimodal biosignal models, but many additional models could be included in future comparisons. Although chronological splits prevent temporal leakage and patient overlap, all data originate from a single healthcare system. Generalizability across institutions, acquisition devices, and patient populations therefore remains to be evaluated. Finally, while the task suite spans 20 clinically relevant outcomes, it does not cover the full spectrum of cardiovascular and systemic conditions, nor does it assess prospective clinical integration. Future work should also explore more advanced multimodal fusion strategies and temporal aggregation methods, and evaluate robustness under distribution shifts and real-world deployment settings.

\emph{The main actionable takeaways from our study are:}
(1) Domain-specific biosignal FMs should be preferred over general time-series FMs for cardiovascular prediction tasks.
(2) Multimodal ECG and PPG data should be leveraged whenever available, as even simple late-fusion strategies yield consistent improvements over unimodal inputs.
(3) Longer signal segments should be prioritized, as increased temporal coverage (up to the full 10-minute signal) consistently improves performance across models and tasks.
(4) Larger FM variants should not be assumed to outperform smaller ones without task-specific validation.
(5) Hand-crafted ECG domain features should be included as a simple yet strong baseline to verify that FMs provide meaningful gains, and can offer complementary benefits when combined with learned FM representations.







\section*{Methods}

We describe the construction of the SignalMC-MED benchmark, the downstream prediction tasks, and the unified evaluation framework used to compare FMs. As illustrated in Figure~\ref{fig:method}, each visit comprises a synchronized 10-minute single-lead ECG and PPG signal that is segmented, encoded using frozen FMs, aggregated into visit-level representations, and evaluated using linear models across 20 clinically relevant tasks. The following sections detail the dataset construction, preprocessing, task definitions, evaluation protocol, and the evaluated FMs and baselines.

\subsubsection*{Dataset Construction}
SignalMC-MED is derived from the MC-MED dataset~\cite{kansal2025mc, PhysioNet-mc-med-1.0.1}, which consists of data from 118{,}385 emergency department visits involving 70{,}545 unique adult patients, collected at Stanford Health Care between 2020 and 2022. MC-MED includes continuously acquired single-lead ECG and PPG recordings captured using bedside patient monitors, together with structured clinical information such as visit outcomes, vital signs, prior medical history, and laboratory test results. Among the 54{,}024 visits with recordings of both ECG and PPG, we identify 24{,}567 visits with at least 10 minutes of temporally overlapping ECG and PPG data within the first recorded signal segment. After signal preprocessing and quality control, this yields 22{,}256 visits with valid synchronized ECG and PPG signals, corresponding to 17{,}841 unique patients. For each visit, a contiguous 10-minute ECG/PPG segment is extracted and used for all downstream analyses.

We adopt the chronological train/val/test splits provided by MC-MED to reflect a realistic clinical deployment scenario. All \textit{val} visits occur after the final \textit{train} visit, and all \textit{test} visits occur after the final \textit{val} visit, with no patient appearing in more than one split. This gives 15{,}832 \textit{train} visits (13{,}011 unique patients), 2{,}030 \textit{val} visits (1{,}882 unique patients), and 1{,}995 \textit{test} visits (1{,}870 unique patients).

\paragraph{Signal Preprocessing \& Segmentation}
Each 10-minute ECG and PPG signal undergoes standardized preprocessing consisting of resampling to a common frequency, signal cleaning using NeuroKit2~\cite{neurokit2}, and Z-score normalization. ECG and PPG signals are resampled to the same frequency, matched to the pretraining sampling frequency expected by each model (Table~\ref{table:fms}), with 250\,Hz used as the default where applicable. The preprocessed 10-minute signals are then divided into non-overlapping 10-second segments, which form the basic input units for all FMs. For experiments involving shorter signal durations, only the initial contiguous subset of segments corresponding to the desired duration is used. As signal quality control, hand-crafted domain features are also extracted from both the ECG and PPG signals (see \emph{Domain Features} below for details). Visits for which any preprocessing step fails are excluded from the analysis.

\subsubsection*{Benchmark Tasks}
SignalMC-MED includes 20 downstream prediction tasks spanning both regression and binary classification. Regression tasks include patient age and eight laboratory measurements obtained during the emergency department visit. Classification tasks include patient sex, emergency department disposition, and the presence of specific ICD-10 diagnoses in the patient's prior medical history, grouped into nine clinically meaningful code sets. These tasks were selected to capture a diverse range of physiological, demographic, and clinically relevant outcomes. Table~\ref{table:csfm-base_results_all-20-tasks_val} summarizes all 20 benchmark tasks and reports CSFM-base performance across them. The positive class prevalence for each of the binary classification tasks across the \textit{train}, \textit{val}, and \textit{test} splits is reported in Table~\ref{table:cls_tasks}. These results provide a reference for relative task difficulty and highlight the heterogeneity of signal predictability across clinical outcomes.

\paragraph{Demographics}
We include two demographic tasks: age regression and sex classification. Age is treated as a continuous target representing patient age in years at the time of the visit, and sex is formulated as a binary classification task using the recorded biological sex. Labels for both tasks are extracted from the \texttt{visits.csv} file in MC-MED.

\paragraph{ED Disposition}
Emergency department disposition is formulated as a binary classification task distinguishing visits resulting in discharge from those resulting in inpatient admission, observation status, or intensive care unit admission. Disposition labels are extracted from \texttt{visits.csv}.

\paragraph{Laboratory Measurements}
Eight laboratory regression tasks corresponding to commonly ordered laboratory tests are included (e.g., electrolytes, renal function markers, and metabolic indicators). For each task, the laboratory value measured during the visit is used as the regression target, extracted from \texttt{labs.csv}. To ensure unambiguous supervision, only visits with exactly one recorded measurement of the corresponding laboratory test are included. The resulting number of \textit{train}, \textit{val}, and \textit{test} visits therefore varies slightly by task, as reported in Table~\ref{table:labs_reg_visits}.

\paragraph{ICD-10 Codes}
Nine binary classification tasks are defined for detecting whether a patient has a documented history of specific ICD-10 diagnoses \emph{prior to} the emergency department visit. Diagnoses are grouped into clinically meaningful categories, including atrial fibrillation, heart failure, cardiac device implantation, chronic kidney disease, diabetes mellitus, sleep-related breathing disorder, anemia, respiratory compromise, and venous thromboembolism. Labels are derived from the patient prior medical history in \texttt{pmh.csv}.

\begin{figure*}[t]
    \centering
    \includegraphics[width=1.0\textwidth]{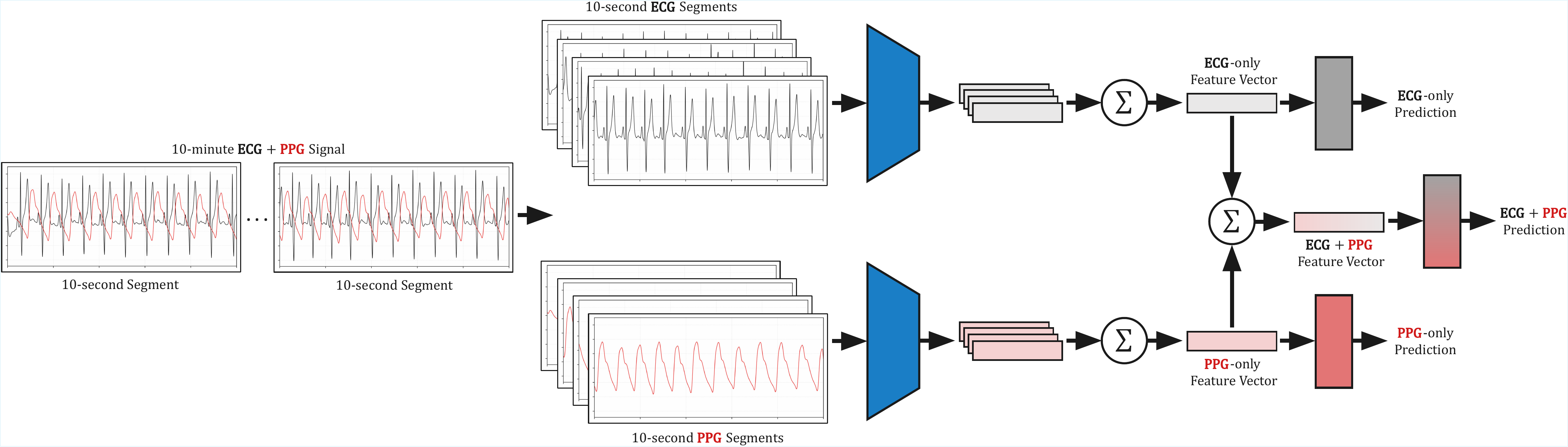}
    \caption{\textbf{Detailed evaluation framework for SignalMC-MED.} A 10-minute synchronized single-lead ECG (\textbf{black}) and PPG (\textcolor{OrangeRed2}{\textbf{red}}) signal is divided into non-overlapping 10-second segments. ECG and PPG segments are processed separately by a frozen FM (\textcolor{Cerulean}{\textbf{blue}}) to extract a feature vector for each 10-second segment. The evaluated FM is applied to both ECG and PPG inputs, regardless of its pretraining modality. The resulting segment-level features are aggregated via mean pooling ($\Sigma$) to form ECG-only and PPG-only visit-level feature vectors, which are then averaged to obtain an ECG + PPG feature vector of the same dimensionality. Separate linear models are trained on the ECG-only, PPG-only, and ECG + PPG representations to generate task-specific predictions across the 20 benchmark tasks.}
    \label{fig:method}
\end{figure*}

\subsubsection*{Evaluation Framework}
All FMs are evaluated using a unified linear probing framework designed to isolate representation quality, as illustrated in Figure~\ref{fig:method}. Models are used exclusively as frozen feature extractors, and no fine-tuning is performed on downstream tasks.

\paragraph{Feature Extraction \& Aggregation}
Each 10-minute signal is divided into non-overlapping 10-second segments. For each segment, a fixed-dimensional feature vector is extracted from the FM. Segment-level features are aggregated by computing the mean feature vector across all segments within the selected signal duration (i.e., averaging all 60 segment feature vectors for the full 10-minute signal, or averaging the first 6 feature vectors for a 1-minute signal). For the multimodal setting, simple feature-level fusion is performed by averaging the ECG-only and PPG-only visit-level feature vectors. For each visit and FM, this yields an ECG-only, a PPG-only, and an ECG + PPG feature vector of the same dimensionality.

\paragraph{Downstream Prediction Models}
Downstream prediction is performed using linear models on top of the aggregated visit-level feature vectors. Ridge regression is used for regression tasks and logistic regression for classification tasks. To minimize task-specific tuning, a single regularization hyperparameter is selected per FM. The ridge penalty is optimized on the \textit{val} split using the age regression task, and the logistic regression inverse regularization strength is optimized using the sex classification task. These hyperparameters are then fixed and applied across all remaining tasks. Hyperparameter selection is performed separately for ECG-only, PPG-only, and ECG + PPG representations, as well as for each evaluated signal duration. Apart from this regularization selection, all downstream training settings are held constant across modalities, signal lengths, and FMs to ensure fair comparison. The implementation uses scikit-learn~\cite{JMLR:v12:pedregosa11a} (version 1.7.2) with default settings. Feature vectors are standardized using the mean and standard deviation computed on the \textit{train} split, and the same transformation is applied to both \textit{val} and \textit{test}.



\paragraph{Evaluation Metrics \& Reporting}
Model performance is evaluated using Pearson correlation for regression tasks and AUROC for classification tasks. To assess data efficiency and robustness, downstream models are trained using 10\%, 25\%, 50\%, and 100\% of the available \textit{train} visits. For each percentage, training visits are sampled with replacement from the \textit{train} split (including the 100\% setting), and both downstream training and hyperparameter selection are repeated five times. Results across all 20 benchmark tasks (Table~\ref{table:csfm-base_results_all-20-tasks_val}, Table~\ref{table:results_all-20-tasks_val_ecg_std}~-~\ref{table:results_all-20-tasks_val_ecg_ppg_mean_std_10}) are reported as mean $\pm$ standard deviation over the five resampling repetitions, for a given train visit percentage.

For summary comparisons (Table~\ref{table:main_results_ecg-baseline-features-ppg-baseline-features-20sec_test}), the 20 benchmark tasks are grouped into five categories: age regression, sex classification, emergency department disposition classification, laboratory value regression (aggregated across eight laboratory tests), and ICD-10 diagnosis classification (aggregated across nine diagnosis groups). For each resampling repetition, performance is first aggregated across tasks within each category and across train visit percentages, yielding one aggregate value per category. The reported results correspond to the mean $\pm$ standard deviation across the five resampling repetitions. This produces five aggregate mean performance values per FM and modality. Based on these, models are ranked separately for ECG-only, PPG-only, and ECG + PPG by computing ranks within each of the five task categories and then averaging these ranks across categories (Table~\ref{table:main_results_ecg-baseline-features-ppg-baseline-features-20sec_test_rank_subsets}). The mean model ranks are used to facilitate cross-model comparison while reducing sensitivity to any individual task.

For figure-based comparisons (Figure~\ref{figure:main_results_ecg-baseline-features-ppg-baseline-features-20sec_test}~-~\ref{figure:results_ecg-baseline-features-ppg-baseline-features-20sec_val_domain-features-concat_comp}, Figure~\ref{figure:results_ecg-baseline-features-ppg-baseline-features-20sec_val_fusion_part2}~-~\ref{figure:results_ecg-baseline-features-ppg-baseline-features-20sec_val_csfm-base_domain-features_comp}), classification performance is aggregated across the eleven binary classification tasks and regression performance across the nine regression tasks. For clarity of visualization, figures display only the mean performance across the five resampling repetitions, omitting standard deviation bars to avoid visual clutter.

\subsubsection*{Evaluated Models}
We evaluate eight representative approaches spanning general time-series FMs, ECG-specific and PPG-specific biosignal FMs, a multimodal ECG + PPG FM, and hand-crafted domain features. An overview of the evaluated FMs is provided in Table~\ref{table:fms}, including model size, pretraining data, feature dimensionality, and expected input sampling frequency. None of the evaluated FMs were pretrained on MC-MED, ensuring strict separation between model pretraining and benchmark evaluation. When multiple model sizes are available within a family, the variant used for primary comparisons is selected based on validation performance (Table~\ref{table:main_results_ecg-baseline-features-ppg-baseline-features-20sec_all-models_val}).

\paragraph{General Time-Series FMs}
We evaluate MOMENT~\citep{goswami2024moment} and Chronos-Bolt~\citep{ansari2024chronos}, which are modality-agnostic time-series FMs pretrained on large-scale heterogeneous time-series corpora. Multiple model sizes are considered (Table~\ref{table:fms}). Because these models were pretrained on data with varying sampling frequencies and do not specify an expected input frequency, we resample ECG and PPG signals to the default 250\,Hz across all experiments. For Chronos-Bolt, segment-level feature vectors are extracted by using the \textit{ChronosBoltPipeline.embed} function and mean pooling across all patch token embeddings. For MOMENT, we use the \textit{MOMENTPipeline} in its default embedding configuration.

\paragraph{Unimodal Biosignal FMs}
We evaluate four unimodal biosignal FMs: three ECG-specific models (D-BETA~\citep{hung2025d-beta}, ECGFounder~\citep{li2025ecgfounder} and xECG~\citep{lunelli2025xecg}) and one PPG-specific model (PaPaGei~\citep{pillai2025papagei}). The ECG-specific models are pretrained on large-scale ECG datasets and incorporate architectural and training strategies tailored to electrocardiographic morphology and rhythm modeling. PaPaGei is pretrained on large-scale PPG datasets comprising millions of short PPG segments. We use the PaPaGei-S variant across all experiments. Inputs are resampled to match each model’s expected sampling frequency according to Table~\ref{table:fms}. Because xECG is based on the xLSTM architecture~\citep{beck2024xlstm}, which supports long input sequences, we additionally evaluate a variant that directly processes the entire 10-minute signal and extracts a single visit-level feature vector. This long-input variant is denoted \textit{xECG-10min}. Although all these models are pretrained exclusively on single-modality data, we evaluate them on both ECG-only, PPG-only, and ECG + PPG inputs to assess cross-modality generalization.

For D-BETA, segment-level feature vectors are extracted using the \textit{get\_ecg\_feats} function, where the ECG/PPG segment is provided as the second channel and the remaining 11 channels are set to zero. For ECGFounder, we use the single-lead model version and extract features using \textit{ft\_1lead\_ECGFounder} with \textit{return\_features} set to true. For both xECG and xECG-10min, the ECG/PPG segment is provided as the second channel with the remaining leads set to zero. For PaPaGei, we use \textit{ResNet1DMoE} configured with a single input channel.

\paragraph{Multimodal Biosignal FM}
We evaluate CSFM~\citep{gu2025csfm}, a multimodal biosignal FM pretrained on more than 2 million ECGs and over 250{,}000 paired ECG and PPG recordings. CSFM is designed to learn both shared and modality-specific representations across cardiovascular signals. Inputs are resampled to the 250\,Hz sampling frequency used during pretraining, and multiple model sizes are considered (Table~\ref{table:fms}). CSFM supports ECG-only, PPG-only, and joint ECG + PPG inputs via an internal multimodal fusion mechanism. For consistency with the other evaluated models, primary results are reported using simple feature-level fusion obtained by averaging the unimodal representations. A direct comparison between this feature-level fusion and CSFM's internal fusion mechanism is provided in Figure~\ref{figure:results_ecg-baseline-features-ppg-baseline-features-20sec_val_csfm-fusion}. For ECG-only and PPG-only inputs, the respective segment is provided as a single-channel input with $\mathrm{channel} = [1]$ (ECG) or $\mathrm{channel} = [12]$ (PPG). For joint ECG + PPG inputs, the two segments are concatenated and provided as a two-channel input with $\mathrm{channel} = [1, 12]$.

\paragraph{Domain Features}
As a simple yet strong baseline, we evaluate hand-crafted domain features extracted independently from ECG and PPG signals. Following CSFM~\citep{gu2025csfm}, we compute a 54-dimensional ECG feature vector using NeuroKit2~\cite{neurokit2} and a 306-dimensional PPG feature vector using pyPPG~\citep{goda2024pyppg}. Signals are resampled to 250\,Hz prior to feature extraction for consistency. Features are extracted at the segment level and aggregated using the same mean pooling strategy applied to FM representations. Standard 10-second segments are used for ECG, whereas 20-second segments are used for PPG to satisfy pyPPG's minimum input length requirement. For PPG, we also consider 60-second and 120-second segments (Table~\ref{table:main_results_ecg-baseline-features-ppg-baseline-features-20sec_all-models_val}). In the multimodal ECG + PPG setting, ECG and PPG domain feature vectors are concatenated to form a combined 360-dimensional representation.




\subsubsection*{Ethics Statement}
This study exclusively utilizes publicly available, de-identified data from the MC-MED dataset~\cite{kansal2025mc, PhysioNet-mc-med-1.0.1}, and complies with the data use policies governing MC-MED.

\section*{Data Availability}

SignalMC-MED is derived from the MC-MED dataset which is available on PhysioNet~\cite{goldberger2000physiobank} at \url{https://physionet.org/content/mc-med}.
\section*{Code Availability}

The code for preprocessing, feature extraction and model evaluation is available at \url{https://github.com/fregu856/SignalMC-MED}. Further implementation details are available from FKG upon reasonable request.
\section*{Acknowledgments}

DAC was funded by an NIHR Research Professorship; a Royal Academy of Engineering Research Chair; and the InnoHK Hong Kong Centre for Cerebro-cardiovascular Engineering (COCHE); and was supported by the National Institute for Health and Care Research (NIHR) Oxford Biomedical Research Centre (BRC) and the Pandemic Sciences Institute at the University of Oxford. XG was supported by funding from the Engineering and Physical Sciences Research Council IAA (EP/X525777/1), the Medical Research Council (MR/X50273X/1), and a Smart Data Research UK Fellowship (UKRI4004). This study was also supported by the NIHR Health Protection Research Unit in Healthcare Associated Infections and Antimicrobial Resistance (NIHR207397), a partnership between the UK Health Security Agency (UKHSA), the University of Oxford, and the NIHR Oxford Biomedical Research Centre. The views expressed are those of the authors and not necessarily those of the NIHR, UKHSA or the Department of Health and Social Care.
\section*{Author Contributions}

FKG was responsible for project conceptualization, software implementation, preparation of figures and tables, and manuscript drafting. XG, MC, and PP contributed to the design of experiments and to the interpretation and presentation of results. DAC and XG supervised the project. DAC and DWE acquired funding. All authors contributed to manuscript revision and finalization.

\section*{Competing Interests}

The authors declare no competing interests.
\section*{Declaration of Generative AI Use}

During the preparation of this manuscript, the authors used ChatGPT 5.2 to assist with language editing and drafting. All content produced using this tool was critically reviewed, edited and validated by the authors, who take full responsibility for the final content of the manuscript.

{
    \small
    \bibliographystyle{ieeenat_fullname}
    \bibliography{references}
}

\clearpage
\appendix
\onecolumn

\renewcommand{\thefigure}{S\arabic{figure}}
\setcounter{figure}{0}

\renewcommand{\thetable}{S\arabic{table}}
\setcounter{table}{0}

\renewcommand{\theequation}{S\arabic{equation}}
\setcounter{equation}{0}

\subsection*{\centering{SignalMC-MED: A Multimodal Benchmark for Evaluating Biosignal\\ Foundation Models on Single-Lead ECG and PPG}}
\section*{\centering{Supplementary Material}}

\vspace{6.0mm}

\section{Supplementary Figures}
\label{appendix:figures}

\begin{figure*}[t]
    \centering
    \includegraphics[width=1.0\textwidth]{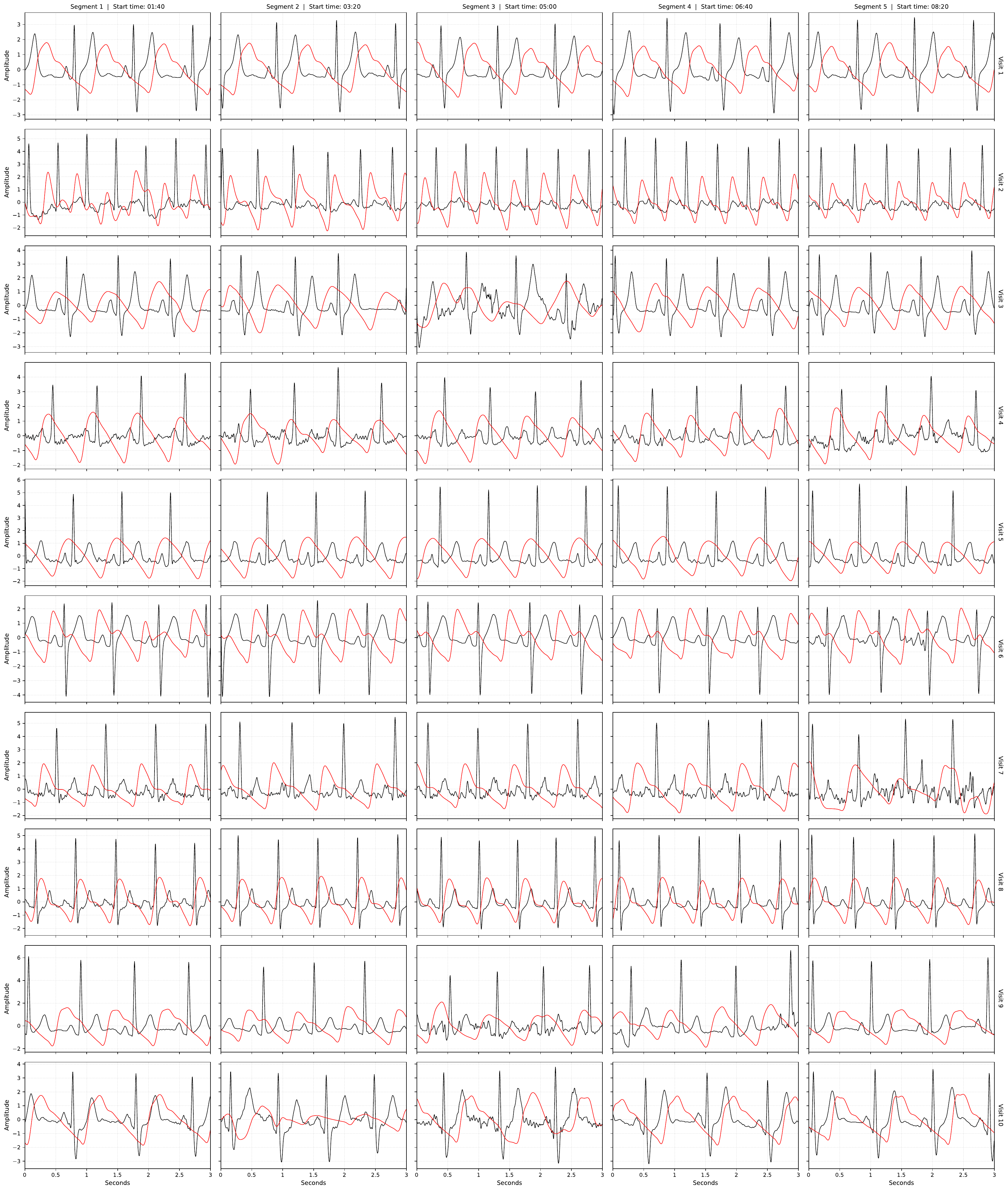}
    \caption{\textbf{Additional synchronized ECG and PPG signals in SignalMC-MED}. An extension of Figure~\ref{fig:synced_ecg_ppg_signal_examples}, with seven more examples of synchronized single-lead ECG (\textbf{black}) and PPG (\textcolor{OrangeRed2}{\textbf{red}}) signals from visits in the \textit{train} set.}
    \label{fig:synced_ecg_ppg_signal_examples_more}
\end{figure*}

\begin{figure*}[t]
\centering
    \begin{subfigure}[t]{0.33\textwidth}
        \centering%
        \includegraphics[clip, trim=1.5cm 0.95cm 1.85cm 0.95cm, width=1.0\linewidth]{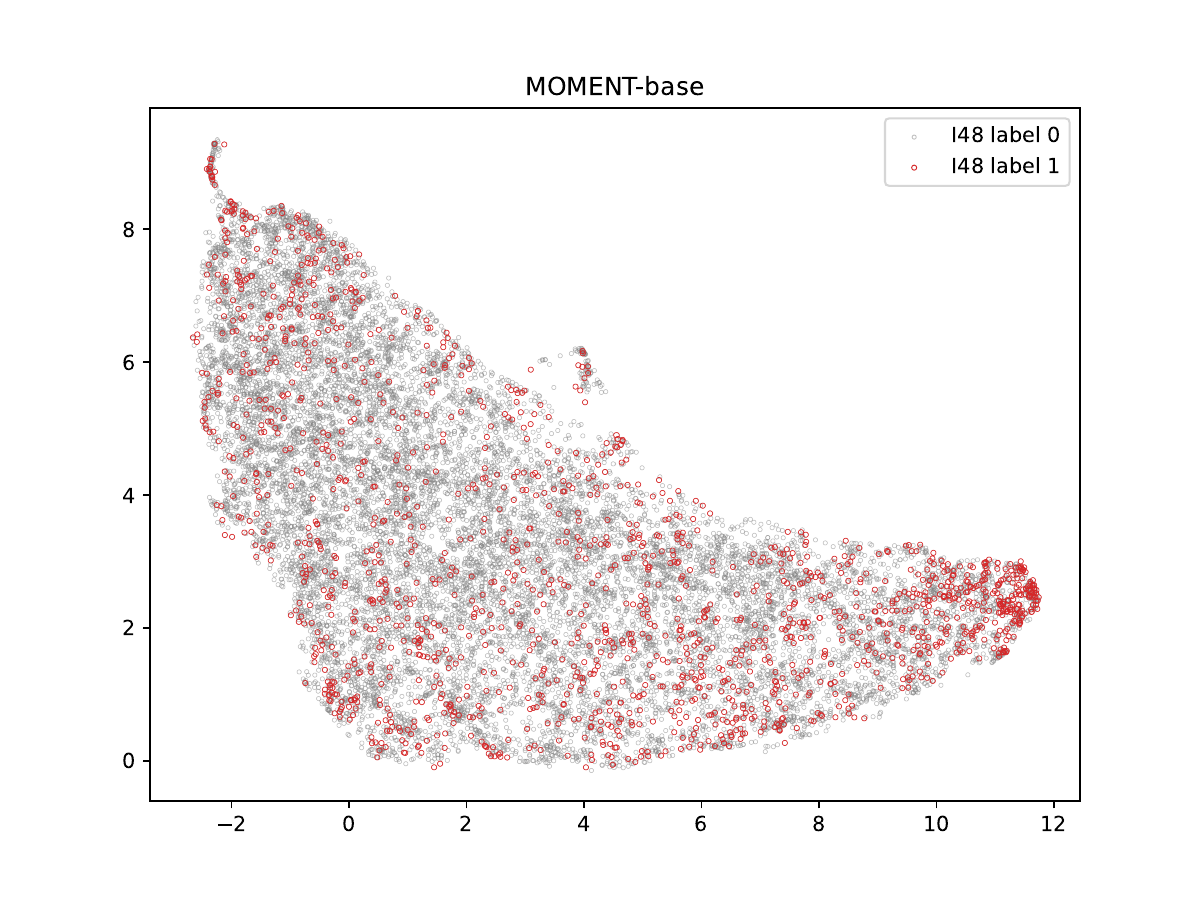}
    \end{subfigure}
    \begin{subfigure}[t]{0.33\textwidth}
        \centering%
        \includegraphics[clip, trim=1.5cm 0.95cm 1.85cm 0.95cm, width=1.0\linewidth]{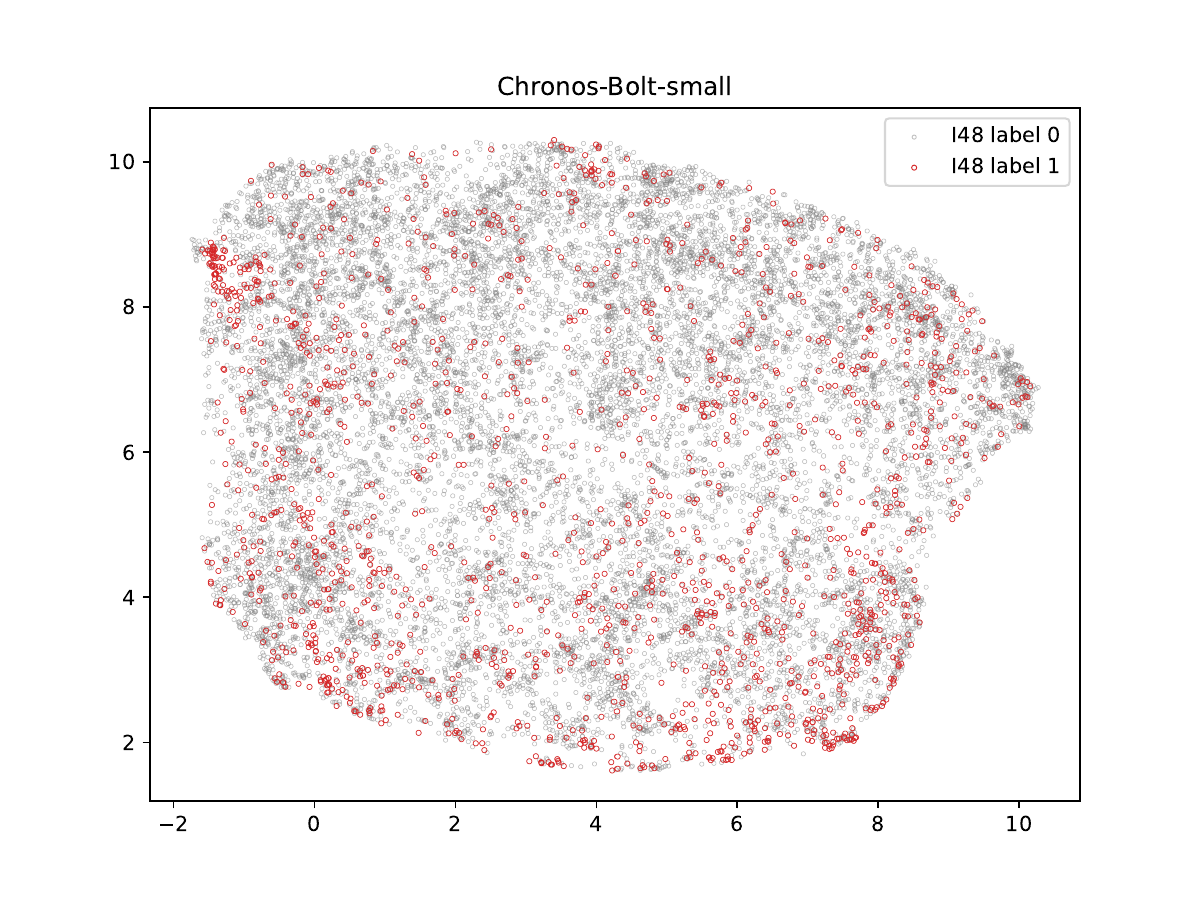}
    \end{subfigure}
    \begin{subfigure}[t]{0.33\textwidth}
        \centering%
        \includegraphics[clip, trim=1.5cm 0.95cm 1.85cm 0.95cm, width=1.0\linewidth]{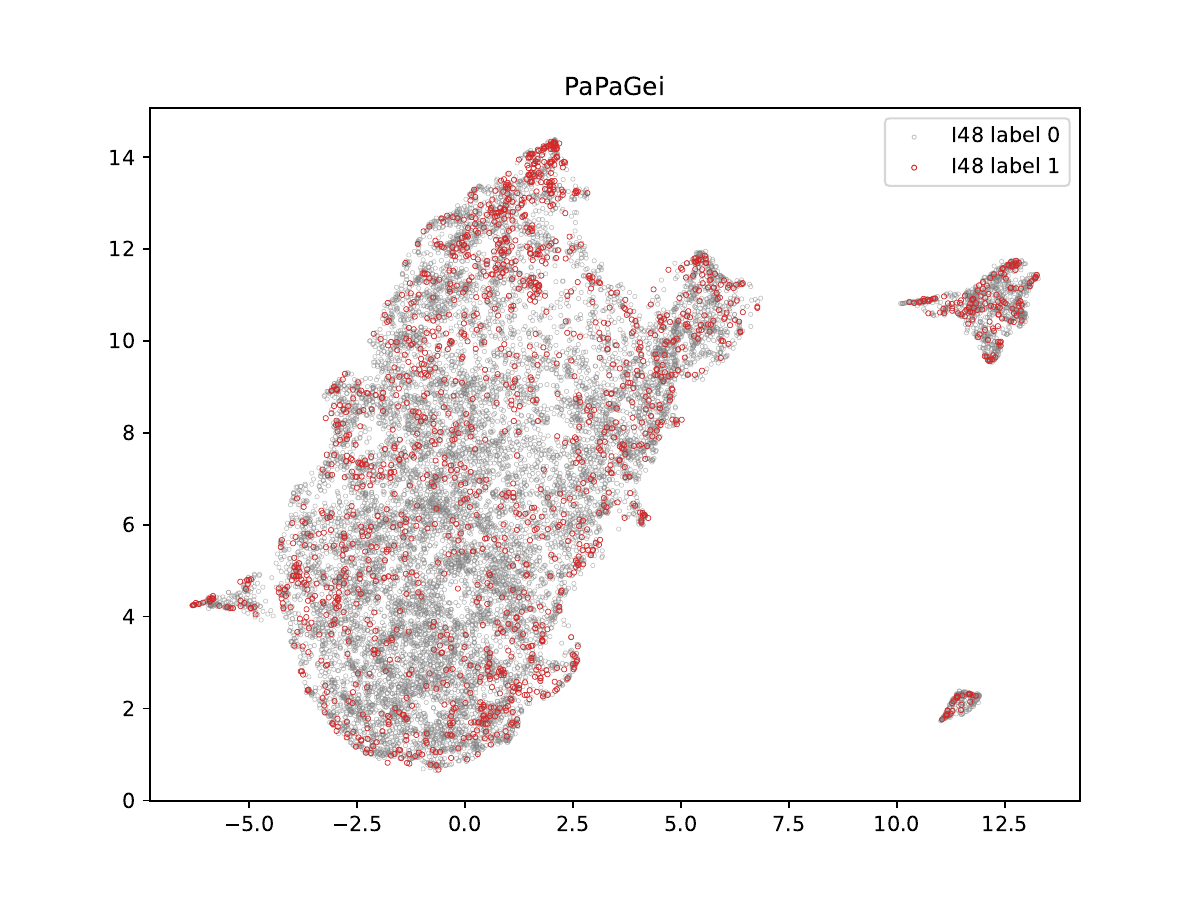}
    \end{subfigure}
    \begin{subfigure}[t]{0.33\textwidth}
        \centering%
        \includegraphics[clip, trim=1.5cm 0.95cm 1.85cm 0.95cm, width=1.0\linewidth]{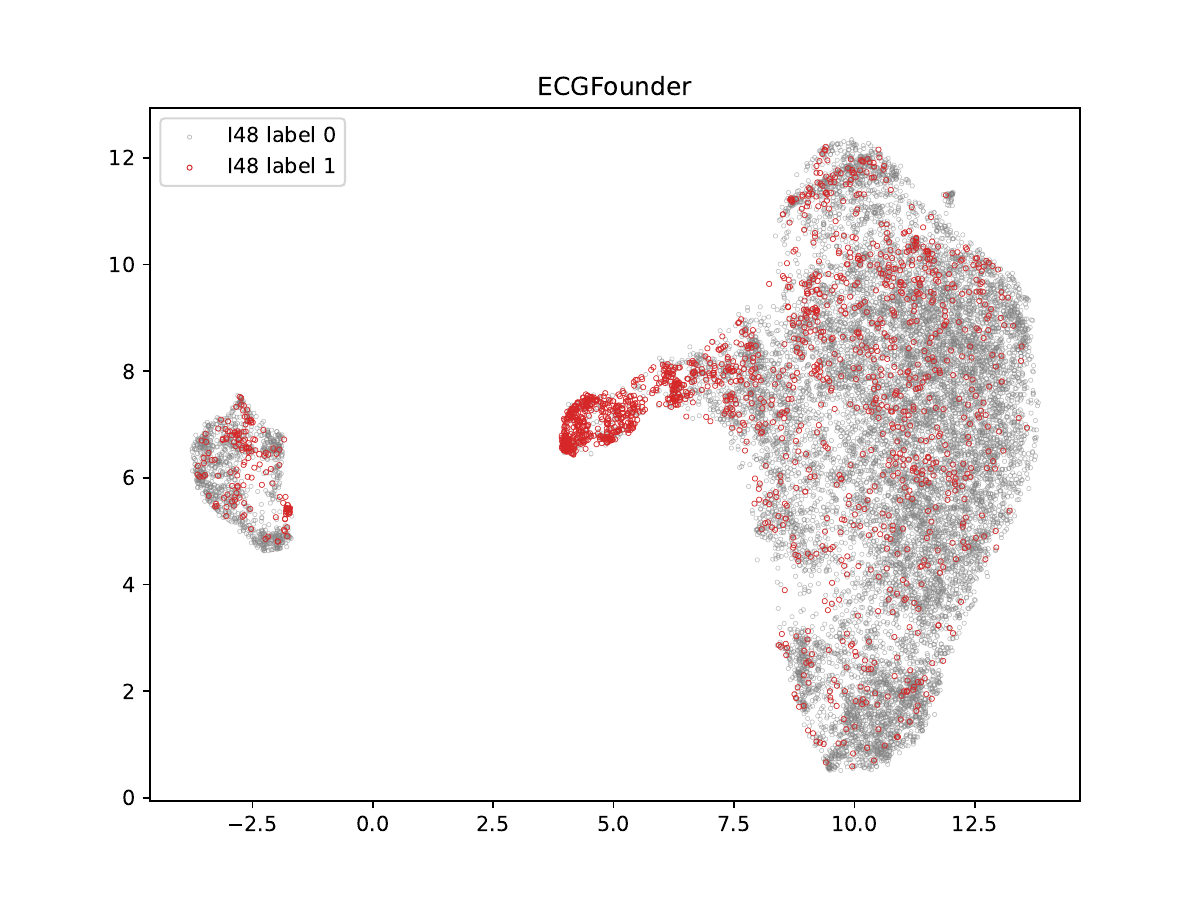}
    \end{subfigure}
    \begin{subfigure}[t]{0.33\textwidth}
        \centering%
        \includegraphics[clip, trim=1.5cm 0.95cm 1.85cm 0.95cm, width=1.0\linewidth]{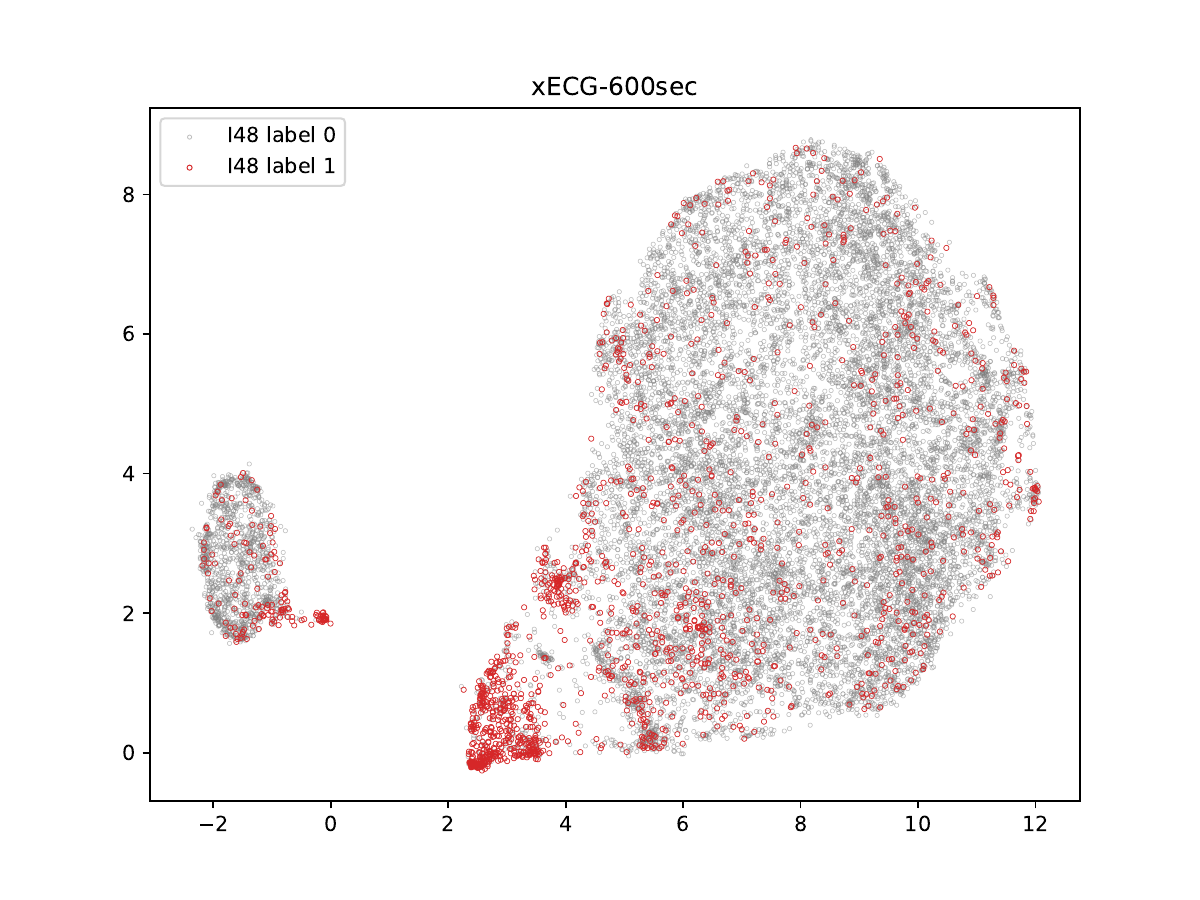}
    \end{subfigure}
    \begin{subfigure}[t]{0.33\textwidth}
        \centering%
        \includegraphics[clip, trim=1.5cm 0.95cm 1.85cm 0.95cm, width=1.0\linewidth]{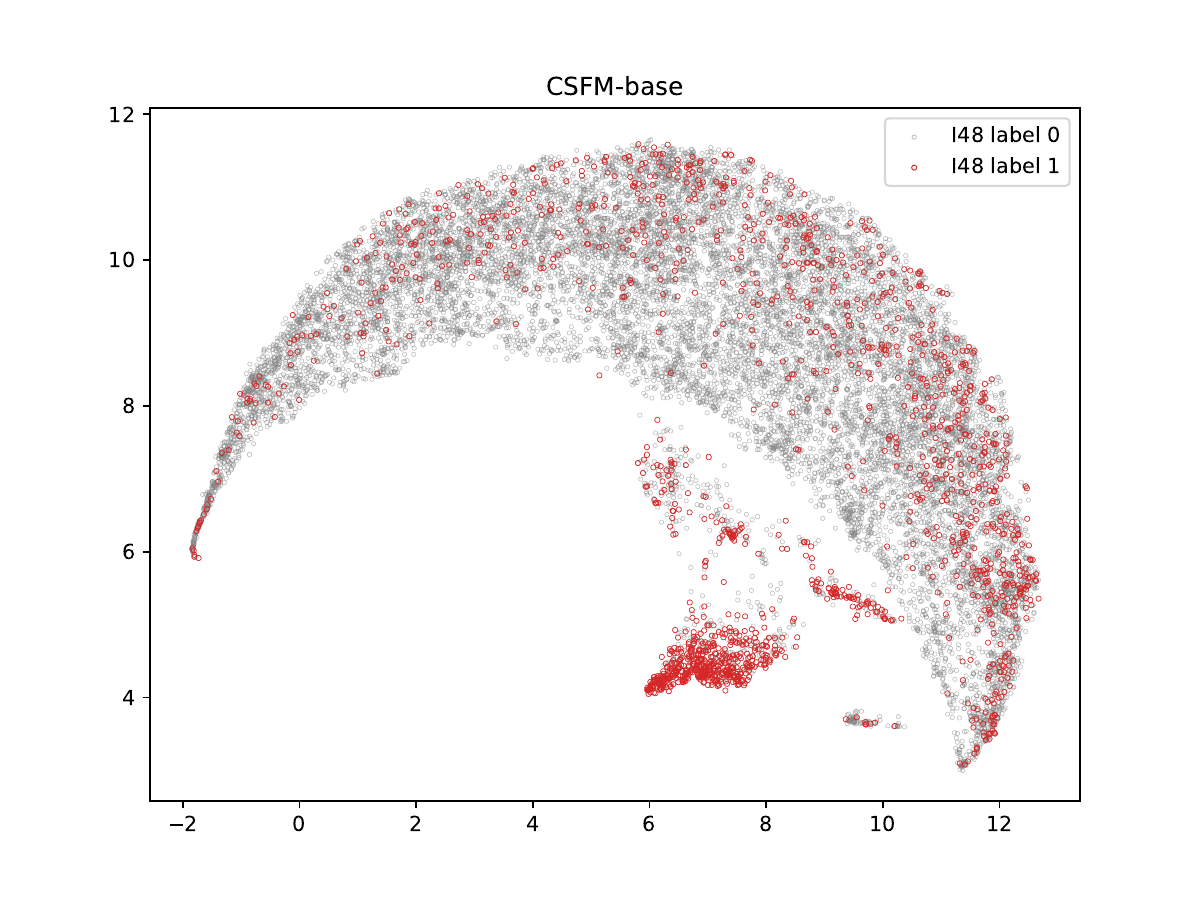}
    \end{subfigure}
  \caption{\textbf{UMAP visualizations colored by atrial fibrillation (I48) label}. UMAP projections of the \textit{train} set \textit{ECG + PPG} feature vectors for six selected models. CSFM-base, xECG-10min and ECGFounder exhibit clearer separation between positive and negative cases compared to general time-series FMs (MOMENT-base, Chronos-Bolt-small) and PaPaGei.}
  \label{fig:umap_plots_I48}
\end{figure*}

This section contains Figure~\ref{fig:synced_ecg_ppg_signal_examples_more} - \ref{fig:umap_plots_I48}.

\input{figures/results_ecg-baseline-features-ppg-baseline-features-20sec_val_fusion_part2}

\input{figures/results_ecg-baseline-features-ppg-baseline-features-20sec_val_signal-lens_part2}

\input{figures/results_ecg-baseline-features-ppg-baseline-features-20sec_val_signal-lens_domain-features}

\input{figures/results_ecg-baseline-features-ppg-baseline-features-20sec_val_fusion-signal-lens_part2}

\input{figures/results_ecg-baseline-features-ppg-baseline-features-20sec_val_model-sizes}
\input{figures/results_ecg-baseline-features-ppg-baseline-features-20sec_val_model-sizes_pca-256}

\input{figures/results_ecg-baseline-features-ppg-baseline-features-20sec_val_domain-features-concat_comp_ppg}

\input{figures/results_ecg-baseline-features-ppg-baseline-features-20sec_val_csfm-fusion}

\input{figures/results_ecg-baseline-features-ppg-baseline-features-20sec_val_csfm-base_domain-features_comp}

\clearpage
\section{Supplementary Tables}
\label{appendix:tables}

This section contains Table~\ref{table:main_results_ecg-baseline-features-ppg-baseline-features-20sec_all-models_val} - \ref{table:labs_reg_visits}.

\begin{table*}[t]
	\caption{\textbf{Comparison of all model variants on the \textit{val} set}, with results averaged across training with 10\%, 25\%, 50\%, and 100\% of the train visits (five resampling repeats per train fraction). Regression tasks are evaluated using Pearson correlation ($\uparrow$), and classification tasks using AUROC ($\uparrow$). \textcolor{OrangeRed2}{\textbf{Bold red}} marks the best result in each column, \textbf{bold} marks the second-best, and \underline{underline} marks the third-best. Based on these results and the corresponding ranking in Table~\ref{table:main_results_ecg-baseline-features-ppg-baseline-features-20sec_all-models_val_rank}, one variant each for MOMENT, Chronos-Bolt, xECG, CSFM, and the PPG domain features is selected for inclusion in the main model comparison on the \textit{test} set. These selected variants are here indicated with italics and a superscript $^\ast$ next to the model name.}\vspace{-2.0mm}	
    \label{table:main_results_ecg-baseline-features-ppg-baseline-features-20sec_all-models_val}
    \centering
	\resizebox{0.95\linewidth}{!}{%
		\begin{tabular}{ll@{\hspace{0.8cm}}ccccc}
\toprule
\textbf{Model Name} &\textbf{Modality} &\textbf{Age reg} &\textbf{Sex cls} &\textbf{ED\_dispo cls} &\textbf{Labs reg} &\textbf{ICD-10 cls}\\
\midrule
MOMENT-small &ECG-only              &0.668    &0.750    &0.696    &0.239    &0.685\\
MOMENT-small &PPG-only              &0.676    &0.721    &0.665    &0.224    &0.652\\
MOMENT-small &ECG + PPG             &0.747    &0.781    &0.706    &0.272    &0.701\\
\midrule
\textit{MOMENT-base}$^\ast$ &ECG-only              &0.680    &0.757    &0.695    &0.253    &0.683\\
\textit{MOMENT-base}$^\ast$ &PPG-only              &0.691    &0.725    &0.680    &0.233    &0.660\\
\textit{MOMENT-base}$^\ast$ &ECG + PPG             &0.762    &0.791    &0.703    &0.284    &0.700\\
\midrule
MOMENT-large &ECG-only              &0.635    &0.720    &0.687    &0.223    &0.673\\
\bottomrule
\end{tabular}
	}
\end{table*}

\begin{table*}[t] 
	\caption{\textbf{Ranking of all model variants on the \textit{val} set}, computed jointly across ECG-only, PPG-only, and ECG + PPG. Mean model rank ($\downarrow$) across the five aggregated task categories based on Table~\ref{table:main_results_ecg-baseline-features-ppg-baseline-features-20sec_all-models_val}. Based on this ranking, one variant each for MOMENT, Chronos-Bolt, xECG, CSFM, and the PPG domain features is selected for inclusion in the main model comparison on the \textit{test} set.}\vspace{-2.0mm}	
    \label{table:main_results_ecg-baseline-features-ppg-baseline-features-20sec_all-models_val_rank}
    \centering
	\resizebox{0.95\textwidth}{!}{%
		\begin{tabular}{cll@{\hspace{0.8cm}}cl}
\toprule
\textbf{Rank} &\textbf{Model Name} &\textbf{Modality} &\textbf{Model Ranks} &\textbf{Mean Model Rank ($\downarrow$)}\\
\midrule
1    &CSFM-base &ECG + PPG        &1,4,1,1,1        &\hspace{13.0mm}1.6\\
2    &CSFM-large &ECG + PPG        &2,1,4,3,5        &\hspace{13.0mm}3.0\\
3    &xECG-10min &ECG + PPG        &3,2,3,2,7        &\hspace{13.0mm}3.4\\
4    &xECG &ECG + PPG        &4,3,5,4,8        &\hspace{13.0mm}4.8\\
5    &CSFM-base &ECG-only        &9,9,2,5,2        &\hspace{13.0mm}5.4\\
6    &xECG-10min &ECG-only        &11,6,6,6,10        &\hspace{13.0mm}7.8\\
7    &xECG &ECG-only        &12,8,7,8,12        &\hspace{13.0mm}9.4\\
\bottomrule
\end{tabular}
	}
\end{table*}

\begin{table*}[t] 
	\caption{\textbf{Main model ranking on the \textit{test} set}, computed jointly across ECG-only, PPG-only, and ECG + PPG. Mean model rank ($\downarrow$) across the five aggregated task categories based on Table~\ref{table:main_results_ecg-baseline-features-ppg-baseline-features-20sec_test}. This table complements the modality-specific rankings shown in Table~\ref{table:main_results_ecg-baseline-features-ppg-baseline-features-20sec_test_rank_subsets}.}\vspace{-2.0mm}	
    \label{table:main_results_ecg-baseline-features-ppg-baseline-features-20sec_test_rank}
    \centering
	\resizebox{0.725\textwidth}{!}{%
		\begin{tabular}{cll@{\hspace{0.8cm}}cl}
\toprule
\textbf{Rank} &\textbf{Model Name} &\textbf{Modality} &\textbf{Model Ranks} &\textbf{Mean Model Rank ($\downarrow$)}\\
\midrule
1    &CSFM-base &ECG + PPG        &1,1,1,1,3        &\hspace{13.0mm}1.4\\
2    &xECG-10min &ECG + PPG        &2,2,6,3,6        &\hspace{13.0mm}3.8\\
3    &CSFM-base &ECG-only        &7,5,5,2,4        &\hspace{13.0mm}4.6\\
4    &ECGFounder &ECG + PPG        &3,4,8,5,5        &\hspace{13.0mm}5.0\\
5    &Domain Features (concat)\hspace{6.0mm} &ECG + PPG        &8,7,2,9,2        &\hspace{13.0mm}5.6\\
6    &xECG-1omin &ECG-only        &10,3,7,4,8        &\hspace{13.0mm}6.4\\
7    &CSFM-base &PPG-only        &4,9,3,8,9        &\hspace{13.0mm}6.6\\
\bottomrule
\end{tabular}
	}
\end{table*}

\begin{table*}[t]
	\caption{\textbf{Model comparison of ECG-only \textit{val} performance on all 20 benchmark tasks when using \textit{100\%} of the train visits}. Tasks are grouped into two demographics tasks, ED disposition classification, eight laboratory value regression tasks, and nine prior ICD-10 diagnosis classification tasks. Performance on the \textit{val} set when using \textit{100\%} of the train set visits. Training visit are sampled with replacement and both downstream training and hyperparameter selection are repeated five times, results are mean $\pm$ std across repetitions. Regression tasks are evaluated using Pearson correlation ($\uparrow$), and classification tasks using AUROC ($\uparrow$). \textcolor{OrangeRed2}{\textbf{Bold red}} marks the best mean value in each row.}\vspace{-2.0mm}	
    \label{table:results_all-20-tasks_val_ecg_std}
    \centering
	\resizebox{1.0\linewidth}{!}{%
        \begin{tabular}{lcccccccc}
\toprule
\textbf{Task} &\textbf{Domain Features} &\textbf{MOMENT-base} &\textbf{Chronos-Bolt-small} &\textbf{PaPaGei} &\textbf{D-BETA} &\textbf{ECGFounder} &\textbf{xECG-10min} &\textbf{CSFM-base}\\
\midrule
Age reg       &0.696 $\pm$ 0.001  &0.694 $\pm$ 0.001  &0.710 $\pm$ 0.001  &0.559 $\pm$ 0.001  &0.688 $\pm$ 0.005  &0.778 $\pm$ 0.003  &\textcolor{OrangeRed2}{\textbf{0.785}} $\pm$ 0.001  &\textcolor{OrangeRed2}{\textbf{0.785}} $\pm$ 0.002\\
Sex cls       &0.676 $\pm$ 0.001  &0.774 $\pm$ 0.002  &0.770 $\pm$ 0.003  &0.675 $\pm$ 0.002  &0.730 $\pm$ 0.003  &0.788 $\pm$ 0.003  &\textcolor{OrangeRed2}{\textbf{0.822}} $\pm$ 0.002  &0.816 $\pm$ 0.002\\
\midrule
ED\_dispo cls &0.729 $\pm$ 0.001  &0.703 $\pm$ 0.003  &0.712 $\pm$ 0.005  &0.655 $\pm$ 0.003  &0.695 $\pm$ 0.004  &0.725 $\pm$ 0.003  &\textcolor{OrangeRed2}{\textbf{0.745}} $\pm$ 0.001  &0.744 $\pm$ 0.001\\
\midrule
Potassium reg  &0.268 $\pm$ 0.004  &0.322 $\pm$ 0.008  &0.304 $\pm$ 0.010  &0.194 $\pm$ 0.016  &0.188 $\pm$ 0.023  &0.315 $\pm$ 0.005  &0.383 $\pm$ 0.007  &\textcolor{OrangeRed2}{\textbf{0.386}} $\pm$ 0.007\\
Calcium reg    &0.155 $\pm$ 0.011  &0.150 $\pm$ 0.007  &0.127 $\pm$ 0.018  &0.077 $\pm$ 0.012  &0.098 $\pm$ 0.016  &0.155 $\pm$ 0.014  &0.205 $\pm$ 0.010  &\textcolor{OrangeRed2}{\textbf{0.209}} $\pm$ 0.010\\
eGFR reg       &0.546 $\pm$ 0.001  &0.494 $\pm$ 0.007  &0.509 $\pm$ 0.005  &0.369 $\pm$ 0.006  &0.481 $\pm$ 0.015  &0.579 $\pm$ 0.007  &0.585 $\pm$ 0.003  &\textcolor{OrangeRed2}{\textbf{0.591}} $\pm$ 0.006\\
Glucose reg    &\textcolor{OrangeRed2}{\textbf{0.297}} $\pm$ 0.007  &0.177 $\pm$ 0.010  &0.199 $\pm$ 0.008  &0.157 $\pm$ 0.013  &0.150 $\pm$ 0.018  &0.244 $\pm$ 0.009  &0.257 $\pm$ 0.014  &0.262 $\pm$ 0.017\\
\bottomrule
\end{tabular}
	}
\end{table*}
\begin{table*}[t]
	\caption{\textbf{Model comparison of PPG-only \textit{val} performance on all 20 benchmark tasks when using \textit{100\%} of the train visits}. Identical format and content as Table~\ref{table:results_all-20-tasks_val_ecg_std}, but for \textit{PPG-only}. Performance on the \textit{val} set when using \textit{100\%} of the train set visits.}\vspace{-2.0mm}	
    \label{table:results_all-20-tasks_val_ppg_std}
    \centering
	\resizebox{1.0\linewidth}{!}{%
        \begin{tabular}{lcccccccc}
\toprule
\textbf{Task} &\textbf{Domain Features} &\textbf{MOMENT-base} &\textbf{Chronos-Bolt-small} &\textbf{PaPaGei} &\textbf{D-BETA} &\textbf{ECGFounder} &\textbf{xECG-10min} &\textbf{CSFM-base}\\
\midrule
Age reg       &0.734 $\pm$ 0.003  &0.704 $\pm$ 0.002  &0.726 $\pm$ 0.002  &0.673 $\pm$ 0.003  &0.663 $\pm$ 0.007  &0.703 $\pm$ 0.005  &0.778 $\pm$ 0.002  &\textcolor{OrangeRed2}{\textbf{0.799}} $\pm$ 0.002\\
Sex cls       &0.744 $\pm$ 0.004  &0.731 $\pm$ 0.004  &0.733 $\pm$ 0.005  &0.691 $\pm$ 0.002  &0.715 $\pm$ 0.004  &0.735 $\pm$ 0.004  &\textcolor{OrangeRed2}{\textbf{0.767}} $\pm$ 0.003  &\textcolor{OrangeRed2}{\textbf{0.767}} $\pm$ 0.003\\
\midrule
ED\_dispo cls &0.706 $\pm$ 0.003  &0.688 $\pm$ 0.005  &0.689 $\pm$ 0.001  &0.673 $\pm$ 0.005  &0.671 $\pm$ 0.006  &0.686 $\pm$ 0.003  &0.707 $\pm$ 0.004  &\textcolor{OrangeRed2}{\textbf{0.726}} $\pm$ 0.001\\
\midrule
Potassium reg  &0.152 $\pm$ 0.016  &0.175 $\pm$ 0.016  &0.178 $\pm$ 0.008  &0.164 $\pm$ 0.014  &0.115 $\pm$ 0.025  &0.135 $\pm$ 0.025  &0.145 $\pm$ 0.007  &\textcolor{OrangeRed2}{\textbf{0.190}} $\pm$ 0.006\\
Calcium reg    &0.149 $\pm$ 0.010  &0.122 $\pm$ 0.009  &0.156 $\pm$ 0.015  &0.117 $\pm$ 0.012  &0.100 $\pm$ 0.009  &0.121 $\pm$ 0.040  &0.165 $\pm$ 0.016  &\textcolor{OrangeRed2}{\textbf{0.183}} $\pm$ 0.015\\
eGFR reg       &0.485 $\pm$ 0.005  &0.439 $\pm$ 0.004  &0.479 $\pm$ 0.004  &0.436 $\pm$ 0.007  &0.399 $\pm$ 0.014  &0.419 $\pm$ 0.063  &0.513 $\pm$ 0.007  &\textcolor{OrangeRed2}{\textbf{0.567}} $\pm$ 0.005\\
Glucose reg    &0.281 $\pm$ 0.014  &0.255 $\pm$ 0.004  &0.292 $\pm$ 0.013  &0.236 $\pm$ 0.012  &0.193 $\pm$ 0.017  &0.188 $\pm$ 0.062  &0.266 $\pm$ 0.015  &\textcolor{OrangeRed2}{\textbf{0.297}} $\pm$ 0.009\\
\bottomrule
\end{tabular}
	}
\end{table*}
\begin{table*}[t]
	\caption{\textbf{Model comparison of ECG + PPG \textit{val} performance on all 20 benchmark tasks when using \textit{100\%} of the train visits}. Identical format and content as Table~\ref{table:results_all-20-tasks_val_ecg_std}, but for \textit{ECG + PPG}. Performance on the \textit{val} set when using \textit{100\%} of the train set visits.}\vspace{-2.0mm}	
    \label{table:results_all-20-tasks_val_ecg_ppg_mean_std}
    \centering
	\resizebox{1.0\linewidth}{!}{%
        \begin{tabular}{lcccccccc}
\toprule
\textbf{Task} &\textbf{Domain Features} &\textbf{MOMENT-base} &\textbf{Chronos-Bolt-small} &\textbf{PaPaGei} &\textbf{D-BETA} &\textbf{ECGFounder} &\textbf{xECG-10min} &\textbf{CSFM-base}\\
\midrule
Age reg       &0.803 $\pm$ 0.003  &0.776 $\pm$ 0.001  &0.781 $\pm$ 0.002  &0.697 $\pm$ 0.002  &0.747 $\pm$ 0.003  &0.822 $\pm$ 0.002  &0.831 $\pm$ 0.001  &\textcolor{OrangeRed2}{\textbf{0.842}} $\pm$ 0.001\\
Sex cls       &0.771 $\pm$ 0.004  &0.810 $\pm$ 0.001  &0.808 $\pm$ 0.003  &0.728 $\pm$ 0.002  &0.770 $\pm$ 0.001  &0.819 $\pm$ 0.002  &\textcolor{OrangeRed2}{\textbf{0.848}} $\pm$ 0.001  &0.840 $\pm$ 0.002\\
\midrule
ED\_dispo cls &0.734 $\pm$ 0.003  &0.710 $\pm$ 0.006  &0.720 $\pm$ 0.003  &0.676 $\pm$ 0.003  &0.702 $\pm$ 0.005  &0.732 $\pm$ 0.002  &\textcolor{OrangeRed2}{\textbf{0.750}} $\pm$ 0.002  &\textcolor{OrangeRed2}{\textbf{0.750}} $\pm$ 0.002\\
\midrule
Potassium reg  &0.238 $\pm$ 0.017  &0.333 $\pm$ 0.009  &0.326 $\pm$ 0.005  &0.216 $\pm$ 0.010  &0.192 $\pm$ 0.019  &0.308 $\pm$ 0.010  &\textcolor{OrangeRed2}{\textbf{0.389}} $\pm$ 0.005  &0.386 $\pm$ 0.006\\
Calcium reg    &0.203 $\pm$ 0.010  &0.181 $\pm$ 0.011  &0.180 $\pm$ 0.009  &0.096 $\pm$ 0.003  &0.130 $\pm$ 0.021  &0.189 $\pm$ 0.007  &0.231 $\pm$ 0.015  &\textcolor{OrangeRed2}{\textbf{0.265}} $\pm$ 0.010\\
eGFR reg       &0.573 $\pm$ 0.008  &0.515 $\pm$ 0.008  &0.541 $\pm$ 0.007  &0.448 $\pm$ 0.008  &0.488 $\pm$ 0.017  &0.585 $\pm$ 0.006  &0.596 $\pm$ 0.003  &\textcolor{OrangeRed2}{\textbf{0.613}} $\pm$ 0.003\\
Glucose reg    &0.311 $\pm$ 0.014  &0.254 $\pm$ 0.010  &0.279 $\pm$ 0.008  &0.205 $\pm$ 0.010  &0.197 $\pm$ 0.022  &0.295 $\pm$ 0.007  &0.303 $\pm$ 0.010  &\textcolor{OrangeRed2}{\textbf{0.322}} $\pm$ 0.010\\
\bottomrule
\end{tabular}
	}
\end{table*}
\begin{table*}[t]
	\caption{\textbf{Model comparison of ECG-only \textit{val} performance on all 20 benchmark tasks when using \textit{50\%} of the train visits}. Tasks are grouped into two demographics tasks, ED disposition classification, eight laboratory value regression tasks, and nine prior ICD-10 diagnosis classification tasks. Performance on the \textit{val} set when using \textit{50\%} of the train set visits. Training visit are sampled with replacement and both downstream training and hyperparameter selection are repeated five times, results are mean $\pm$ std across repetitions. Regression tasks are evaluated using Pearson correlation ($\uparrow$), and classification tasks using AUROC ($\uparrow$). \textcolor{OrangeRed2}{\textbf{Bold red}} marks the best mean value in each row.}\vspace{-2.0mm}	
    \label{table:results_all-20-tasks_val_ecg_std_50}
    \centering
	\resizebox{1.0\linewidth}{!}{%
        \begin{tabular}{lcccccccc}
\toprule
\textbf{Task} &\textbf{Domain Features} &\textbf{MOMENT-base} &\textbf{Chronos-Bolt-small} &\textbf{PaPaGei} &\textbf{D-BETA} &\textbf{ECGFounder} &\textbf{xECG-10min} &\textbf{CSFM-base}\\
\midrule
Age reg       &0.694 $\pm$ 0.001  &0.685 $\pm$ 0.003  &0.699 $\pm$ 0.004  &0.554 $\pm$ 0.003  &0.672 $\pm$ 0.009  &0.775 $\pm$ 0.002  &0.776 $\pm$ 0.002  &\textcolor{OrangeRed2}{\textbf{0.780}} $\pm$ 0.002\\
Sex cls       &0.674 $\pm$ 0.005  &0.766 $\pm$ 0.006  &0.757 $\pm$ 0.006  &0.673 $\pm$ 0.006  &0.725 $\pm$ 0.002  &0.779 $\pm$ 0.003  &\textcolor{OrangeRed2}{\textbf{0.817}} $\pm$ 0.002  &0.811 $\pm$ 0.006\\
\midrule
ED\_dispo cls &0.728 $\pm$ 0.002  &0.699 $\pm$ 0.008  &0.704 $\pm$ 0.003  &0.653 $\pm$ 0.003  &0.691 $\pm$ 0.005  &0.720 $\pm$ 0.004  &0.733 $\pm$ 0.004  &\textcolor{OrangeRed2}{\textbf{0.737}} $\pm$ 0.006\\
\midrule
Potassium reg  &0.268 $\pm$ 0.009  &0.302 $\pm$ 0.017  &0.279 $\pm$ 0.008  &0.179 $\pm$ 0.005  &0.178 $\pm$ 0.022  &0.294 $\pm$ 0.026  &0.370 $\pm$ 0.010  &\textcolor{OrangeRed2}{\textbf{0.377}} $\pm$ 0.005\\
Calcium reg    &0.150 $\pm$ 0.016  &0.133 $\pm$ 0.016  &0.109 $\pm$ 0.012  &0.088 $\pm$ 0.012  &0.077 $\pm$ 0.025  &0.142 $\pm$ 0.011  &\textcolor{OrangeRed2}{\textbf{0.199}} $\pm$ 0.016  &0.195 $\pm$ 0.005\\
eGFR reg       &0.540 $\pm$ 0.005  &0.493 $\pm$ 0.004  &0.500 $\pm$ 0.006  &0.352 $\pm$ 0.009  &0.459 $\pm$ 0.019  &0.559 $\pm$ 0.007  &\textcolor{OrangeRed2}{\textbf{0.581}} $\pm$ 0.003  &0.580 $\pm$ 0.005\\
Glucose reg    &\textcolor{OrangeRed2}{\textbf{0.297}} $\pm$ 0.002  &0.185 $\pm$ 0.023  &0.188 $\pm$ 0.026  &0.152 $\pm$ 0.013  &0.160 $\pm$ 0.025  &0.233 $\pm$ 0.025  &0.270 $\pm$ 0.011  &0.249 $\pm$ 0.011\\
\bottomrule
\end{tabular}
	}
\end{table*}
\begin{table*}[t]
	\caption{\textbf{Model comparison of PPG-only \textit{val} performance on all 20 benchmark tasks when using \textit{50\%} of the train visits}. Identical format and content as Table~\ref{table:results_all-20-tasks_val_ecg_std_50}, but for \textit{PPG-only}. Performance on the \textit{val} set when using \textit{50\%} of the train set visits.}\vspace{-2.0mm}	
    \label{table:results_all-20-tasks_val_ppg_std_50}
    \centering
	\resizebox{1.0\linewidth}{!}{%
        \begin{tabular}{lcccccccc}
\toprule
\textbf{Task} &\textbf{Domain Features} &\textbf{MOMENT-base} &\textbf{Chronos-Bolt-small} &\textbf{PaPaGei} &\textbf{D-BETA} &\textbf{ECGFounder} &\textbf{xECG-10min} &\textbf{CSFM-base}\\
\midrule
Age reg       &0.718 $\pm$ 0.013  &0.701 $\pm$ 0.004  &0.716 $\pm$ 0.002  &0.674 $\pm$ 0.003  &0.604 $\pm$ 0.068  &0.655 $\pm$ 0.083  &0.768 $\pm$ 0.003  &\textcolor{OrangeRed2}{\textbf{0.793}} $\pm$ 0.002\\
Sex cls       &0.740 $\pm$ 0.004  &0.729 $\pm$ 0.002  &0.726 $\pm$ 0.004  &0.692 $\pm$ 0.001  &0.711 $\pm$ 0.008  &0.734 $\pm$ 0.006  &0.764 $\pm$ 0.006  &\textcolor{OrangeRed2}{\textbf{0.765}} $\pm$ 0.004\\
\midrule
ED\_dispo cls &0.704 $\pm$ 0.004  &0.687 $\pm$ 0.004  &0.682 $\pm$ 0.002  &0.670 $\pm$ 0.004  &0.664 $\pm$ 0.003  &0.677 $\pm$ 0.008  &0.703 $\pm$ 0.002  &\textcolor{OrangeRed2}{\textbf{0.721}} $\pm$ 0.004\\
\midrule
Potassium reg  &0.136 $\pm$ 0.008  &0.173 $\pm$ 0.010  &0.167 $\pm$ 0.018  &0.172 $\pm$ 0.007  &0.095 $\pm$ 0.036  &0.136 $\pm$ 0.019  &0.169 $\pm$ 0.014  &\textcolor{OrangeRed2}{\textbf{0.175}} $\pm$ 0.008\\
Calcium reg    &0.133 $\pm$ 0.018  &0.125 $\pm$ 0.016  &0.157 $\pm$ 0.010  &0.101 $\pm$ 0.006  &0.100 $\pm$ 0.015  &0.119 $\pm$ 0.026  &0.171 $\pm$ 0.015  &\textcolor{OrangeRed2}{\textbf{0.188}} $\pm$ 0.018\\
eGFR reg       &0.467 $\pm$ 0.009  &0.425 $\pm$ 0.005  &0.458 $\pm$ 0.006  &0.424 $\pm$ 0.002  &0.322 $\pm$ 0.020  &0.381 $\pm$ 0.041  &0.503 $\pm$ 0.005  &\textcolor{OrangeRed2}{\textbf{0.552}} $\pm$ 0.006\\
Glucose reg    &0.245 $\pm$ 0.042  &0.242 $\pm$ 0.020  &0.258 $\pm$ 0.029  &0.245 $\pm$ 0.008  &0.160 $\pm$ 0.025  &0.220 $\pm$ 0.034  &\textcolor{OrangeRed2}{\textbf{0.273}} $\pm$ 0.012  &0.272 $\pm$ 0.007\\
\bottomrule
\end{tabular}
	}
\end{table*}
\begin{table*}[t]
	\caption{\textbf{Model comparison of ECG + PPG \textit{val} performance on all 20 benchmark tasks when using \textit{50\%} of the train visits}. Identical format and content as Table~\ref{table:results_all-20-tasks_val_ecg_std_50}, but for \textit{ECG + PPG}. Performance on the \textit{val} set when using \textit{50\%} of the train set visits.}\vspace{-2.0mm}	
    \label{table:results_all-20-tasks_val_ecg_ppg_mean_std_50}
    \centering
	\resizebox{1.0\linewidth}{!}{%
        \begin{tabular}{lcccccccc}
\toprule
\textbf{Task} &\textbf{Domain Features} &\textbf{MOMENT-base} &\textbf{Chronos-Bolt-small} &\textbf{PaPaGei} &\textbf{D-BETA} &\textbf{ECGFounder} &\textbf{xECG-10min} &\textbf{CSFM-base}\\
\midrule
Age reg       &0.797 $\pm$ 0.003  &0.768 $\pm$ 0.003  &0.774 $\pm$ 0.002  &0.687 $\pm$ 0.005  &0.730 $\pm$ 0.011  &0.818 $\pm$ 0.003  &0.823 $\pm$ 0.002  &\textcolor{OrangeRed2}{\textbf{0.839}} $\pm$ 0.002\\
Sex cls       &0.767 $\pm$ 0.003  &0.798 $\pm$ 0.004  &0.798 $\pm$ 0.005  &0.727 $\pm$ 0.004  &0.765 $\pm$ 0.002  &0.815 $\pm$ 0.003  &\textcolor{OrangeRed2}{\textbf{0.842}} $\pm$ 0.002  &0.835 $\pm$ 0.002\\
\midrule
ED\_dispo cls &0.731 $\pm$ 0.005  &0.708 $\pm$ 0.004  &0.715 $\pm$ 0.005  &0.673 $\pm$ 0.005  &0.695 $\pm$ 0.005  &0.724 $\pm$ 0.003  &0.736 $\pm$ 0.004  &\textcolor{OrangeRed2}{\textbf{0.745}} $\pm$ 0.006\\
\midrule
Potassium reg  &0.223 $\pm$ 0.017  &0.315 $\pm$ 0.012  &0.297 $\pm$ 0.007  &0.199 $\pm$ 0.007  &0.189 $\pm$ 0.012  &0.269 $\pm$ 0.035  &\textcolor{OrangeRed2}{\textbf{0.384}} $\pm$ 0.012  &0.376 $\pm$ 0.006\\
Calcium reg    &0.192 $\pm$ 0.020  &0.165 $\pm$ 0.017  &0.165 $\pm$ 0.014  &0.104 $\pm$ 0.015  &0.100 $\pm$ 0.022  &0.164 $\pm$ 0.008  &0.220 $\pm$ 0.017  &\textcolor{OrangeRed2}{\textbf{0.254}} $\pm$ 0.010\\
eGFR reg       &0.568 $\pm$ 0.004  &0.508 $\pm$ 0.008  &0.529 $\pm$ 0.004  &0.423 $\pm$ 0.011  &0.467 $\pm$ 0.025  &0.564 $\pm$ 0.008  &0.588 $\pm$ 0.003  &\textcolor{OrangeRed2}{\textbf{0.598}} $\pm$ 0.004\\
Glucose reg    &0.298 $\pm$ 0.029  &0.254 $\pm$ 0.020  &0.264 $\pm$ 0.016  &0.202 $\pm$ 0.011  &0.218 $\pm$ 0.035  &0.281 $\pm$ 0.014  &\textcolor{OrangeRed2}{\textbf{0.313}} $\pm$ 0.014  &0.300 $\pm$ 0.009\\
\bottomrule
\end{tabular}
	}
\end{table*}
\begin{table*}[t]
	\caption{\textbf{Model comparison of ECG-only \textit{val} performance on all 20 benchmark tasks when using \textit{25\%} of the train visits}. Tasks are grouped into two demographics tasks, ED disposition classification, eight laboratory value regression tasks, and nine prior ICD-10 diagnosis classification tasks. Performance on the \textit{val} set when using \textit{25\%} of the train set visits. Training visit are sampled with replacement and both downstream training and hyperparameter selection are repeated five times, results are mean $\pm$ std across repetitions. Regression tasks are evaluated using Pearson correlation ($\uparrow$), and classification tasks using AUROC ($\uparrow$). \textcolor{OrangeRed2}{\textbf{Bold red}} marks the best mean value in each row.}\vspace{-2.0mm}	
    \label{table:results_all-20-tasks_val_ecg_std_25}
    \centering
	\resizebox{1.0\linewidth}{!}{%
        \begin{tabular}{lcccccccc}
\toprule
\textbf{Task} &\textbf{Domain Features} &\textbf{MOMENT-base} &\textbf{Chronos-Bolt-small} &\textbf{PaPaGei} &\textbf{D-BETA} &\textbf{ECGFounder} &\textbf{xECG-10min} &\textbf{CSFM-base}\\
\midrule
Age reg       &0.689 $\pm$ 0.004  &0.677 $\pm$ 0.005  &0.681 $\pm$ 0.003  &0.538 $\pm$ 0.007  &0.645 $\pm$ 0.026  &0.762 $\pm$ 0.013  &0.769 $\pm$ 0.002  &\textcolor{OrangeRed2}{\textbf{0.775}} $\pm$ 0.004\\
Sex cls       &0.665 $\pm$ 0.006  &0.753 $\pm$ 0.003  &0.737 $\pm$ 0.004  &0.668 $\pm$ 0.005  &0.713 $\pm$ 0.004  &0.764 $\pm$ 0.004  &\textcolor{OrangeRed2}{\textbf{0.806}} $\pm$ 0.003  &0.802 $\pm$ 0.004\\
\midrule
ED\_dispo cls &0.725 $\pm$ 0.002  &0.691 $\pm$ 0.004  &0.690 $\pm$ 0.007  &0.644 $\pm$ 0.003  &0.693 $\pm$ 0.003  &0.711 $\pm$ 0.006  &0.724 $\pm$ 0.002  &\textcolor{OrangeRed2}{\textbf{0.730}} $\pm$ 0.005\\
\midrule
Potassium reg  &0.239 $\pm$ 0.018  &0.291 $\pm$ 0.012  &0.262 $\pm$ 0.018  &0.175 $\pm$ 0.025  &0.187 $\pm$ 0.020  &0.285 $\pm$ 0.008  &0.345 $\pm$ 0.014  &\textcolor{OrangeRed2}{\textbf{0.362}} $\pm$ 0.009\\
Calcium reg    &0.141 $\pm$ 0.016  &0.111 $\pm$ 0.011  &0.101 $\pm$ 0.014  &0.069 $\pm$ 0.010  &0.050 $\pm$ 0.039  &0.133 $\pm$ 0.013  &0.183 $\pm$ 0.008  &\textcolor{OrangeRed2}{\textbf{0.189}} $\pm$ 0.012\\
eGFR reg       &0.536 $\pm$ 0.005  &0.468 $\pm$ 0.005  &0.474 $\pm$ 0.010  &0.352 $\pm$ 0.010  &0.467 $\pm$ 0.021  &0.554 $\pm$ 0.014  &0.558 $\pm$ 0.007  &\textcolor{OrangeRed2}{\textbf{0.567}} $\pm$ 0.005\\
Glucose reg    &\textcolor{OrangeRed2}{\textbf{0.278}} $\pm$ 0.019  &0.161 $\pm$ 0.023  &0.154 $\pm$ 0.019  &0.128 $\pm$ 0.028  &0.148 $\pm$ 0.036  &0.183 $\pm$ 0.044  &0.235 $\pm$ 0.021  &0.241 $\pm$ 0.021\\
\bottomrule
\end{tabular}
	}
\end{table*}
\begin{table*}[t]
	\caption{\textbf{Model comparison of PPG-only \textit{val} performance on all 20 benchmark tasks when using \textit{25\%} of the train visits}. Identical format and content as Table~\ref{table:results_all-20-tasks_val_ecg_std_25}, but for \textit{PPG-only}. Performance on the \textit{val} set when using \textit{25\%} of the train set visits.}\vspace{-2.0mm}	
    \label{table:results_all-20-tasks_val_ppg_std_25}
    \centering
	\resizebox{1.0\linewidth}{!}{%
        \begin{tabular}{lcccccccc}
\toprule
\textbf{Task} &\textbf{Domain Features} &\textbf{MOMENT-base} &\textbf{Chronos-Bolt-small} &\textbf{PaPaGei} &\textbf{D-BETA} &\textbf{ECGFounder} &\textbf{xECG-10min} &\textbf{CSFM-base}\\
\midrule
Age reg       &0.711 $\pm$ 0.007  &0.690 $\pm$ 0.003  &0.704 $\pm$ 0.003  &0.661 $\pm$ 0.003  &0.580 $\pm$ 0.082  &0.696 $\pm$ 0.015  &0.754 $\pm$ 0.003  &\textcolor{OrangeRed2}{\textbf{0.783}} $\pm$ 0.004\\
Sex cls       &0.729 $\pm$ 0.005  &0.724 $\pm$ 0.004  &0.717 $\pm$ 0.004  &0.688 $\pm$ 0.003  &0.699 $\pm$ 0.006  &0.723 $\pm$ 0.004  &0.754 $\pm$ 0.004  &\textcolor{OrangeRed2}{\textbf{0.757}} $\pm$ 0.003\\
\midrule
ED\_dispo cls &0.689 $\pm$ 0.007  &0.678 $\pm$ 0.005  &0.676 $\pm$ 0.007  &0.651 $\pm$ 0.008  &0.660 $\pm$ 0.007  &0.674 $\pm$ 0.009  &0.696 $\pm$ 0.005  &\textcolor{OrangeRed2}{\textbf{0.720}} $\pm$ 0.002\\
\midrule
Potassium reg  &0.092 $\pm$ 0.036  &\textcolor{OrangeRed2}{\textbf{0.156}} $\pm$ 0.018  &0.144 $\pm$ 0.018  &0.148 $\pm$ 0.008  &0.088 $\pm$ 0.024  &0.099 $\pm$ 0.026  &0.146 $\pm$ 0.022  &0.154 $\pm$ 0.019\\
Calcium reg    &0.107 $\pm$ 0.021  &0.106 $\pm$ 0.010  &0.130 $\pm$ 0.017  &0.081 $\pm$ 0.020  &0.085 $\pm$ 0.018  &0.079 $\pm$ 0.033  &0.168 $\pm$ 0.017  &\textcolor{OrangeRed2}{\textbf{0.188}} $\pm$ 0.027\\
eGFR reg       &0.412 $\pm$ 0.021  &0.401 $\pm$ 0.008  &0.422 $\pm$ 0.013  &0.419 $\pm$ 0.006  &0.318 $\pm$ 0.027  &0.407 $\pm$ 0.015  &0.476 $\pm$ 0.008  &\textcolor{OrangeRed2}{\textbf{0.539}} $\pm$ 0.002\\
Glucose reg    &0.204 $\pm$ 0.056  &0.216 $\pm$ 0.020  &0.245 $\pm$ 0.034  &0.234 $\pm$ 0.011  &0.170 $\pm$ 0.034  &0.224 $\pm$ 0.021  &0.239 $\pm$ 0.016  &\textcolor{OrangeRed2}{\textbf{0.256}} $\pm$ 0.022\\
\bottomrule
\end{tabular}
	}
\end{table*}
\begin{table*}[t]
	\caption{\textbf{Model comparison of ECG + PPG \textit{val} performance on all 20 benchmark tasks when using \textit{25\%} of the train visits}. Identical format and content as Table~\ref{table:results_all-20-tasks_val_ecg_std_25}, but for \textit{ECG + PPG}. Performance on the \textit{val} set when using \textit{25\%} of the train set visits.}\vspace{-2.0mm}	
    \label{table:results_all-20-tasks_val_ecg_ppg_mean_std_25}
    \centering
	\resizebox{1.0\linewidth}{!}{%
        \begin{tabular}{lcccccccc}
\toprule
\textbf{Task} &\textbf{Domain Features} &\textbf{MOMENT-base} &\textbf{Chronos-Bolt-small} &\textbf{PaPaGei} &\textbf{D-BETA} &\textbf{ECGFounder} &\textbf{xECG-10min} &\textbf{CSFM-base}\\
\midrule
Age reg       &0.789 $\pm$ 0.003  &0.759 $\pm$ 0.004  &0.762 $\pm$ 0.003  &0.669 $\pm$ 0.002  &0.696 $\pm$ 0.026  &0.798 $\pm$ 0.028  &0.816 $\pm$ 0.002  &\textcolor{OrangeRed2}{\textbf{0.833}} $\pm$ 0.002\\
Sex cls       &0.757 $\pm$ 0.006  &0.785 $\pm$ 0.002  &0.781 $\pm$ 0.002  &0.720 $\pm$ 0.004  &0.748 $\pm$ 0.004  &0.800 $\pm$ 0.003  &\textcolor{OrangeRed2}{\textbf{0.829}} $\pm$ 0.002  &0.826 $\pm$ 0.004\\
\midrule
ED\_dispo cls &0.722 $\pm$ 0.007  &0.697 $\pm$ 0.005  &0.701 $\pm$ 0.004  &0.660 $\pm$ 0.003  &0.697 $\pm$ 0.003  &0.719 $\pm$ 0.005  &0.728 $\pm$ 0.003  &\textcolor{OrangeRed2}{\textbf{0.737}} $\pm$ 0.004\\
\midrule
Potassium reg  &0.157 $\pm$ 0.034  &0.297 $\pm$ 0.007  &0.274 $\pm$ 0.014  &0.187 $\pm$ 0.030  &0.191 $\pm$ 0.029  &0.258 $\pm$ 0.014  &0.360 $\pm$ 0.014  &\textcolor{OrangeRed2}{\textbf{0.361}} $\pm$ 0.004\\
Calcium reg    &0.167 $\pm$ 0.012  &0.141 $\pm$ 0.011  &0.144 $\pm$ 0.023  &0.082 $\pm$ 0.012  &0.074 $\pm$ 0.044  &0.152 $\pm$ 0.012  &0.203 $\pm$ 0.012  &\textcolor{OrangeRed2}{\textbf{0.248}} $\pm$ 0.012\\
eGFR reg       &0.523 $\pm$ 0.014  &0.492 $\pm$ 0.005  &0.508 $\pm$ 0.013  &0.418 $\pm$ 0.007  &0.469 $\pm$ 0.031  &0.549 $\pm$ 0.025  &0.570 $\pm$ 0.004  &\textcolor{OrangeRed2}{\textbf{0.586}} $\pm$ 0.002\\
Glucose reg    &0.243 $\pm$ 0.045  &0.238 $\pm$ 0.018  &0.232 $\pm$ 0.016  &0.180 $\pm$ 0.031  &0.186 $\pm$ 0.040  &0.222 $\pm$ 0.039  &0.283 $\pm$ 0.023  &\textcolor{OrangeRed2}{\textbf{0.288}} $\pm$ 0.016\\
\bottomrule
\end{tabular}
	}
\end{table*}
\begin{table*}[t]
	\caption{\textbf{Model comparison of ECG-only \textit{val} performance on all 20 benchmark tasks when using \textit{10\%} of the train visits}. Tasks are grouped into two demographics tasks, ED disposition classification, eight laboratory value regression tasks, and nine prior ICD-10 diagnosis classification tasks. Performance on the \textit{val} set when using \textit{10\%} of the train set visits. Training visit are sampled with replacement and both downstream training and hyperparameter selection are repeated five times, results are mean $\pm$ std across repetitions. Regression tasks are evaluated using Pearson correlation ($\uparrow$), and classification tasks using AUROC ($\uparrow$). \textcolor{OrangeRed2}{\textbf{Bold red}} marks the best mean value in each row.}\vspace{-2.0mm}	
    \label{table:results_all-20-tasks_val_ecg_std_10}
    \centering
	\resizebox{1.0\linewidth}{!}{%
        \begin{tabular}{lcccccccc}
\toprule
\textbf{Task} &\textbf{Domain Features} &\textbf{MOMENT-base} &\textbf{Chronos-Bolt-small} &\textbf{PaPaGei} &\textbf{D-BETA} &\textbf{ECGFounder} &\textbf{xECG-10min} &\textbf{CSFM-base}\\
\midrule
Age reg       &0.681 $\pm$ 0.003  &0.663 $\pm$ 0.002  &0.657 $\pm$ 0.004  &0.517 $\pm$ 0.008  &0.521 $\pm$ 0.139  &0.726 $\pm$ 0.042  &0.755 $\pm$ 0.004  &\textcolor{OrangeRed2}{\textbf{0.761}} $\pm$ 0.003\\
Sex cls       &0.656 $\pm$ 0.011  &0.735 $\pm$ 0.006  &0.715 $\pm$ 0.004  &0.665 $\pm$ 0.009  &0.706 $\pm$ 0.004  &0.745 $\pm$ 0.007  &\textcolor{OrangeRed2}{\textbf{0.791}} $\pm$ 0.004  &0.788 $\pm$ 0.003\\
\midrule
ED\_dispo cls &0.712 $\pm$ 0.011  &0.687 $\pm$ 0.005  &0.677 $\pm$ 0.014  &0.638 $\pm$ 0.007  &0.690 $\pm$ 0.010  &0.708 $\pm$ 0.015  &0.718 $\pm$ 0.004  &\textcolor{OrangeRed2}{\textbf{0.726}} $\pm$ 0.009\\
\midrule
Potassium reg  &0.220 $\pm$ 0.022  &0.249 $\pm$ 0.012  &0.229 $\pm$ 0.014  &0.141 $\pm$ 0.020  &0.124 $\pm$ 0.064  &0.232 $\pm$ 0.021  &0.310 $\pm$ 0.014  &\textcolor{OrangeRed2}{\textbf{0.329}} $\pm$ 0.011\\
Calcium reg    &0.125 $\pm$ 0.013  &0.094 $\pm$ 0.012  &0.100 $\pm$ 0.019  &0.062 $\pm$ 0.016  &0.060 $\pm$ 0.010  &0.111 $\pm$ 0.005  &\textcolor{OrangeRed2}{\textbf{0.163}} $\pm$ 0.006  &\textcolor{OrangeRed2}{\textbf{0.163}} $\pm$ 0.014\\
eGFR reg       &0.519 $\pm$ 0.010  &0.440 $\pm$ 0.015  &0.440 $\pm$ 0.007  &0.333 $\pm$ 0.016  &0.423 $\pm$ 0.038  &0.483 $\pm$ 0.060  &0.541 $\pm$ 0.005  &\textcolor{OrangeRed2}{\textbf{0.558}} $\pm$ 0.011\\
Glucose reg    &\textcolor{OrangeRed2}{\textbf{0.235}} $\pm$ 0.070  &0.150 $\pm$ 0.028  &0.131 $\pm$ 0.027  &0.117 $\pm$ 0.014  &0.144 $\pm$ 0.057  &0.195 $\pm$ 0.038  &0.211 $\pm$ 0.019  &0.229 $\pm$ 0.025\\
\bottomrule
\end{tabular}
	}
\end{table*}
\begin{table*}[t]
	\caption{\textbf{Model comparison of PPG-only \textit{val} performance on all 20 benchmark tasks when using \textit{10\%} of the train visits}. Identical format and content as Table~\ref{table:results_all-20-tasks_val_ecg_std_10}, but for \textit{PPG-only}. Performance on the \textit{val} set when using \textit{10\%} of the train set visits.}\vspace{-2.0mm}	
    \label{table:results_all-20-tasks_val_ppg_std_10}
    \centering
	\resizebox{1.0\linewidth}{!}{%
        \begin{tabular}{lcccccccc}
\toprule
\textbf{Task} &\textbf{Domain Features} &\textbf{MOMENT-base} &\textbf{Chronos-Bolt-small} &\textbf{PaPaGei} &\textbf{D-BETA} &\textbf{ECGFounder} &\textbf{xECG-10min} &\textbf{CSFM-base}\\
\midrule
Age reg       &0.633 $\pm$ 0.035  &0.668 $\pm$ 0.006  &0.673 $\pm$ 0.006  &0.649 $\pm$ 0.003  &0.549 $\pm$ 0.049  &0.636 $\pm$ 0.023  &0.726 $\pm$ 0.003  &\textcolor{OrangeRed2}{\textbf{0.763}} $\pm$ 0.005\\
Sex cls       &0.716 $\pm$ 0.004  &0.715 $\pm$ 0.005  &0.698 $\pm$ 0.004  &0.682 $\pm$ 0.006  &0.673 $\pm$ 0.013  &0.704 $\pm$ 0.008  &0.734 $\pm$ 0.004  &\textcolor{OrangeRed2}{\textbf{0.744}} $\pm$ 0.006\\
\midrule
ED\_dispo cls &0.676 $\pm$ 0.007  &0.667 $\pm$ 0.006  &0.657 $\pm$ 0.005  &0.632 $\pm$ 0.007  &0.635 $\pm$ 0.008  &0.658 $\pm$ 0.005  &0.672 $\pm$ 0.007  &\textcolor{OrangeRed2}{\textbf{0.712}} $\pm$ 0.005\\
\midrule
Potassium reg  &0.064 $\pm$ 0.023  &0.124 $\pm$ 0.026  &0.122 $\pm$ 0.026  &\textcolor{OrangeRed2}{\textbf{0.129}} $\pm$ 0.020  &0.048 $\pm$ 0.028  &0.103 $\pm$ 0.025  &0.112 $\pm$ 0.020  &0.127 $\pm$ 0.017\\
Calcium reg    &0.076 $\pm$ 0.020  &0.085 $\pm$ 0.027  &0.076 $\pm$ 0.028  &0.072 $\pm$ 0.006  &0.030 $\pm$ 0.010  &0.075 $\pm$ 0.014  &0.118 $\pm$ 0.009  &\textcolor{OrangeRed2}{\textbf{0.160}} $\pm$ 0.013\\
eGFR reg       &0.207 $\pm$ 0.029  &0.374 $\pm$ 0.008  &0.395 $\pm$ 0.019  &0.401 $\pm$ 0.004  &0.214 $\pm$ 0.097  &0.318 $\pm$ 0.082  &0.425 $\pm$ 0.018  &\textcolor{OrangeRed2}{\textbf{0.510}} $\pm$ 0.015\\
Glucose reg    &0.095 $\pm$ 0.061  &0.205 $\pm$ 0.023  &0.202 $\pm$ 0.016  &0.222 $\pm$ 0.019  &0.063 $\pm$ 0.060  &0.226 $\pm$ 0.026  &0.203 $\pm$ 0.026  &\textcolor{OrangeRed2}{\textbf{0.234}} $\pm$ 0.015\\
\bottomrule
\end{tabular}
	}
\end{table*}
\begin{table*}[t]
	\caption{\textbf{Model comparison of ECG + PPG \textit{val} performance on all 20 benchmark tasks when using \textit{10\%} of the train visits}. Identical format and content as Table~\ref{table:results_all-20-tasks_val_ecg_std_10}, but for \textit{ECG + PPG}. Performance on the \textit{val} set when using \textit{10\%} of the train set visits.}\vspace{-2.0mm}	
    \label{table:results_all-20-tasks_val_ecg_ppg_mean_std_10}
    \centering
	\resizebox{1.0\linewidth}{!}{%
        \begin{tabular}{lcccccccc}
\toprule
\textbf{Task} &\textbf{Domain Features} &\textbf{MOMENT-base} &\textbf{Chronos-Bolt-small} &\textbf{PaPaGei} &\textbf{D-BETA} &\textbf{ECGFounder} &\textbf{xECG-10min} &\textbf{CSFM-base}\\
\midrule
Age reg       &0.746 $\pm$ 0.022  &0.747 $\pm$ 0.003  &0.742 $\pm$ 0.005  &0.650 $\pm$ 0.010  &0.528 $\pm$ 0.198  &0.793 $\pm$ 0.004  &0.801 $\pm$ 0.002  &\textcolor{OrangeRed2}{\textbf{0.823}} $\pm$ 0.003\\
Sex cls       &0.746 $\pm$ 0.003  &0.772 $\pm$ 0.006  &0.755 $\pm$ 0.005  &0.714 $\pm$ 0.009  &0.736 $\pm$ 0.007  &0.782 $\pm$ 0.004  &0.813 $\pm$ 0.004  &\textcolor{OrangeRed2}{\textbf{0.814}} $\pm$ 0.004\\
\midrule
ED\_dispo cls &0.712 $\pm$ 0.004  &0.696 $\pm$ 0.007  &0.691 $\pm$ 0.009  &0.652 $\pm$ 0.006  &0.686 $\pm$ 0.015  &0.711 $\pm$ 0.013  &0.718 $\pm$ 0.001  &\textcolor{OrangeRed2}{\textbf{0.728}} $\pm$ 0.006\\
\midrule
Potassium reg  &0.124 $\pm$ 0.053  &0.260 $\pm$ 0.026  &0.226 $\pm$ 0.019  &0.166 $\pm$ 0.021  &0.126 $\pm$ 0.067  &0.199 $\pm$ 0.037  &0.316 $\pm$ 0.027  &\textcolor{OrangeRed2}{\textbf{0.323}} $\pm$ 0.017\\
Calcium reg    &0.113 $\pm$ 0.030  &0.115 $\pm$ 0.012  &0.121 $\pm$ 0.010  &0.075 $\pm$ 0.011  &0.087 $\pm$ 0.009  &0.124 $\pm$ 0.014  &0.180 $\pm$ 0.011  &\textcolor{OrangeRed2}{\textbf{0.210}} $\pm$ 0.007\\
eGFR reg       &0.334 $\pm$ 0.072  &0.469 $\pm$ 0.011  &0.479 $\pm$ 0.008  &0.381 $\pm$ 0.019  &0.422 $\pm$ 0.031  &0.514 $\pm$ 0.037  &0.547 $\pm$ 0.006  &\textcolor{OrangeRed2}{\textbf{0.572}} $\pm$ 0.010\\
Glucose reg    &0.139 $\pm$ 0.068  &0.218 $\pm$ 0.031  &0.193 $\pm$ 0.018  &0.167 $\pm$ 0.008  &0.166 $\pm$ 0.057  &0.202 $\pm$ 0.041  &0.260 $\pm$ 0.015  &\textcolor{OrangeRed2}{\textbf{0.268}} $\pm$ 0.024\\
\bottomrule
\end{tabular}
	}
\end{table*}

\begin{table*}[t]
	\caption{\textbf{Number of \textit{train}, \textit{val}, and \textit{test} visits for each of the eight laboratory regression tasks}. To ensure unambiguous supervision, only visits with exactly one recorded measurement of the corresponding laboratory test are included.}\vspace{-2.0mm}	
    \label{table:labs_reg_visits}
    \centering
	\resizebox{0.60\linewidth}{!}{%
        \begin{tabular}{lcccc}
\toprule
\textbf{Task} &\textbf{Train} &\textbf{Val} &\textbf{Test}\\
\midrule
Potassium reg (``POTASSIUM'')                                &13{,}481    &1{,}700    &1{,}683\\
Calcium reg (``CALCIUM'')                                    &13{,}464    &1{,}700    &1{,}677\\
eGFR reg (``EGFR REFIT WITHOUT RACE (2021)'')\hspace{6.0mm}  &13{,}407    &1{,}694    &1{,}675\\
Glucose reg (``GLUCOSE'')                                    &13{,}100    &1{,}658    &1{,}644\\
Hemoglobin reg (``HEMOGLOBIN (HGB)'')                        &13{,}854    &1{,}772    &1{,}724\\
Albumin reg (``ALBUMIN'')                                    &13{,}538    &1{,}729    &1{,}707\\
BUN reg (``BLOOD UREA NITROGEN (BUN)'')                      &13{,}096    &1{,}658    &1{,}644\\
Sodium reg (``SODIUM'')                                      &13{,}092    &1{,}657    &1{,}643\\
\bottomrule
\end{tabular}
	}
\end{table*}

\begin{table*}[t]
	\caption{\textbf{Positive class prevalence for each of the eleven binary classification tasks}. Percentage of visits labeled as the positive class in the \textit{train}, \textit{val}, and \textit{test} splits. Values are computed at the visit level and are provided to contextualize reported AUROC performance.}\vspace{-2.0mm}	
    \label{table:cls_tasks}
    \centering
	\resizebox{0.85\linewidth}{!}{%
        \begin{tabular}{llccc}
\toprule
\textbf{Task} &\textbf{Label / Description} &\textbf{Train} &\textbf{Val} &\textbf{Test}\\
\midrule
Sex cls &``M'' vs ``F''                                                                   &52.2\%  &52.4\%  &52.5\%\\
\midrule
ED\_dispo cls &``Discharge'' vs \{``Inpatient'', ``Observation'', ``ICU''\}\hspace{6.0mm} &50.5\%  &51.0\%  &51.2\%\\
\midrule
I48 cls &Atrial Fibrillation Detection                           &10.5\%  &8.5\%   &7.6\%\\
I50 cls &Heart Failure Detection                                 &9.8\%   &7.2\%   &6.5\%\\
\{Z95.0, Z95.2, Z95.3\} cls &Cardiac Device Detection            &3.6\%   &2.6\%   &2.4\%\\
N18 cls &Chronic Kidney Disease Detection                        &10.5\%  &7.0\%   &7.4\%\\
\{E10, E11\} cls &Diabetes Mellitus Detection                    &15.4\%  &12.1\%  &11.7\%\\
\{G47.3, E66.2\} cls &Sleep-Related Breathing Disorder Detection &9.3\%   &6.3\%   &7.3\%\\
\{D50, D51, D52, D53, D64\} cls\hspace{5.0mm} &Anemia Detection  &16.1\%  &11.4\%  &11.1\%\\
\{J96, R09.02, R06.0\} cls &Respiratory Compromise Detection     &23.3\%  &15.2\%  &13.1\%\\
\{I26, I80, I82\} cls &Venous Thromboembolism Detection          &8.6\%   &5.4\%   &5.5\%\\
\bottomrule
\end{tabular}
	}
\end{table*}

\end{document}